\documentclass{article}

\usepackage[final]{corl_2022}

\usepackage{microtype}
\usepackage{graphicx}
\usepackage{subfigure}
\usepackage{booktabs} 
\usepackage{algorithm,algorithmic}

\usepackage{amsmath,amssymb,graphicx,color,booktabs,bm,relsize,enumitem,multirow,amsthm,epsfig,caption,mathtools,wrapfig}
\usepackage{setspace}
\usepackage{dsfont}
\usepackage{xspace}
\usepackage[export]{adjustbox}
\newsavebox{\measurebox}

\usepackage[capitalise,noabbrev,nameinlink]{cleveref}
\creflabelformat{equation}{#2\textup{#1}#3}
\creflabelformat{section}{#2\textup{#1}#3}
\creflabelformat{appendix}{#2\textup{#1}#3}
\creflabelformat{figure}{#2\textup{#1}#3}
\crefname{algocf}{Algorithm}{Algorithms}
\Crefname{algocf}{Algorithm}{Algorithms}

\newcommand{\describe}[3][0pt]{\hspace*{.12em}\underbracket[0.5pt][1pt]{#2\hspace*{#1}}_\text{#3}}

\title{Masked World Models for Visual Control}

\author{
  Younggyo Seo$^{1,2,}$\thanks{Work done while visiting UC Berkeley. Correspondence to \texttt{younggyo.seo@kaist.ac.kr}.}
  \quad
  Danijar Hafner$^{2,3,4}$
  \quad
  Hao Liu$^{2}$
  \quad
  Fangchen Liu$^{2}$
  \\[1ex]
  \textbf{Stephen James}$^{2,}$\thanks{Now at Dyson Robot Learning Lab.}
  \quad
  \textbf{Kimin Lee}$^{3}$
  \quad
  \textbf{Pieter Abbeel}$^{2}$
  \\[1ex]
  $^{1}$ KAIST \; $^{2}$ UC Berkeley \; $^{3}$ Google Research \; $^{4}$ University of Toronto
}

\begin{document}
\maketitle

\newcommand{\ALG}{MWM\xspace}

\begin{abstract}
    Visual model-based reinforcement learning (RL) has the potential to enable sample-efficient robot learning from visual observations.
    Yet the current approaches typically train a single model end-to-end for learning both visual representations and dynamics,
    making it difficult to accurately model the interaction between robots and small objects.
    In this work, we introduce a visual model-based RL framework that decouples visual representation learning and dynamics learning.
    Specifically, we train an autoencoder with convolutional layers and vision transformers (ViT) to reconstruct pixels given masked convolutional features, and learn a latent dynamics model that operates on the representations from the autoencoder.
    Moreover, to encode task-relevant information, we introduce an auxiliary reward prediction objective for the autoencoder.
    We continually update both autoencoder and dynamics model using online samples collected from environment interaction.
    We demonstrate that our decoupling approach achieves state-of-the-art performance on a variety of visual robotic tasks from Meta-world and RLBench, \textit{e.g.}, we achieve 81.7\% success rate on 50 visual robotic manipulation tasks from Meta-world, while the baseline achieves 67.9\%. Code is available on the project website: \url{https://sites.google.com/view/mwm-rl}.
\end{abstract}

\section{Introduction}
Model-based reinforcement learning (RL) holds the promise of sample-efficient robot learning by learning a world model and leveraging it for planning~\citep{chua2018deep,deisenroth2011pilco,lenz2015deepmpc} or generating imaginary states for behavior learning~\citep{kurutach2018model,janner2019trust}.
These approaches have also previously been applied to environments with visual observations, by learning an action-conditional video prediction model~\citep{finn2017deep,ebert2018visual} or a latent dynamics model that predicts compact representations in an abstract latent space~\citep{watter2015embed,hafner2019learning}.
However, learning world models on environments with complex visual observations, \textit{e.g.,} accurately modeling interactions with small objects, is an open challenge.

We argue that this difficulty comes from the design of current approaches that typically optimize the world model end-to-end for learning both visual representations and dynamics~\citep{hafner2019learning,zhang2019solar}.
This imposes a trade-off between learning representations and dynamics that can prevent world models from accurately capturing visual details, making it difficult to predict forward into the future.
Another approach is to learn representations and dynamics separately, such as earlier work by \citet{ha2018world} who train a variational autoencoder (VAE)~\citep{kingma2013auto} and a dynamics model on top of the VAE features. 
However, separately-trained VAE representations may not be amenable to dynamics learning~\citep{watter2015embed,zhang2019solar} or may not capture task-relevant details~\citep{ha2018world}.

On the other hand, masked autoencoders~(MAE)~\citep{he2021masked} have recently been proposed as an effective and scalable approach to visual representation learning, by training a self-supervised vision transformer~(ViT)~\citep{dosovitskiy2020image} to reconstruct masked patches.
While it motivates us to learn world models on top of MAE representations, we find that MAE often struggles to capture fine-grained details within patches.
Because capturing visual details, \textit{e.g.,} object positions, is crucial for solving visual control tasks, it is desirable to develop a representation learning method that captures such details but also achieves the benefits of MAE such as stability, compute-efficiency, and scalability.

\begin{figure*} [t!] \centering
    \vspace{-2em}
    \includegraphics[width=0.99\textwidth]{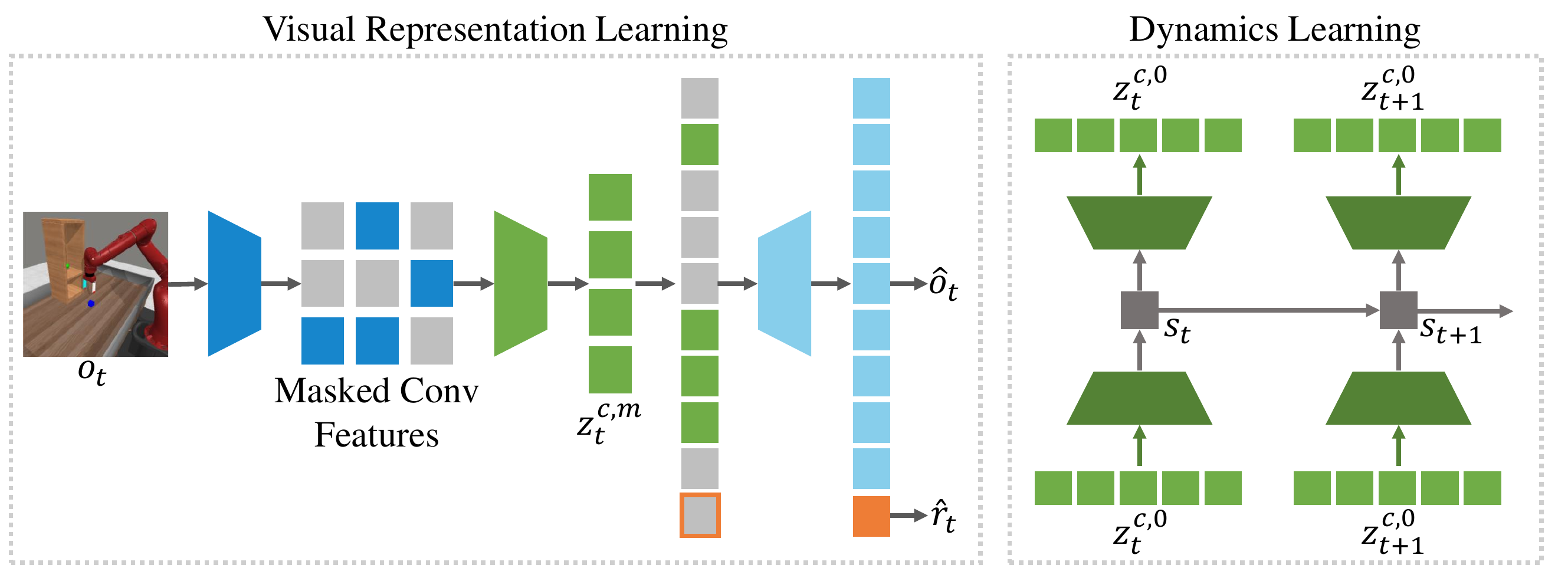}
    \caption{Illustration of our approach. We continually update visual representations and dynamics using online samples collected from environment interaction, by repeating iterative processes of training (Left) an autoencoder with convolutional feature masking and reward prediction and (Right) a latent dynamics model in the latent space of the autoencoder. We note that autoencoder parameters are not updated during dynamics learning.}
    \label{fig:method_overview}
\end{figure*}

In this paper, we present Masked World Models (\ALG), a visual model-based RL algorithm that decouples visual representation learning and dynamics learning.
The key idea of \ALG is to train an autoencoder that reconstructs visual observations with convolutional feature masking, and a latent dynamics model on top of the autoencoder.
By introducing early convolutional layers and masking out convolutional features instead of pixel patches, our approach enables the world model to capture fine-grained visual details from complex visual observations.
Moreover, in order to learn task-relevant information that might not be captured solely by the reconstruction objective,
we introduce an auxiliary reward prediction task for the autoencoder.
Specifically, we separately update visual representations and dynamics by repeating the iterative processes of (i) training the autoencoder with convolutional feature masking and reward prediction, and (ii) learning the latent dynamics model that predicts visual representations from the autoencoder (see~\cref{fig:method_overview}).

\paragraph{Contributions} We highlight the contributions of our paper below:
\begin{itemize}[topsep=1.0pt,itemsep=1.0pt,leftmargin=5.5mm]
    \item [$\bullet$] We demonstrate the effectiveness of decoupling visual representation learning and dynamics learning for visual model-based RL.
    \ALG significantly outperforms a state-of-the-art model-based baseline~\citep{hafner2020mastering} on various visual control tasks from Meta-world~\citep{yu2020meta} and RLBench~\citep{james2020rlbench}.
    \item [$\bullet$] We show that a self-supervised ViT trained to reconstruct visual observations with convolutional feature masking can be effective for visual model-based RL. Interestingly, we find that masking convolutional features can be more effective than pixel patch masking~\citep{he2021masked}, by allowing for capturing fine-grained details within patches.
    This is in contrast to the observation in \citet{touvron2022three}, where both perform similarly on the ImageNet classification task~\citep{deng2009imagenet}.
    \item [$\bullet$] We show that an auxiliary reward prediction task can significantly improve performance by encoding task-relevant information into visual representations.
\end{itemize}

\section{Related Work}
\label{sec:related_work}
\paragraph{World models from visual observations}
There have been several approaches to learn visual representations for model-based approaches via image reconstruction~\citep{finn2017deep,ebert2018visual,watter2015embed,hafner2019learning,zhang2019solar,ha2018world,hafner2020mastering,finn2016deep,hafner2019dream,kaiser2019model}, \textit{e.g.,} learning a video prediction model~\citep{finn2017deep,gupta2022maskvit} or a latent dynamics model~\citep{watter2015embed,hafner2019learning,zhang2019solar}.
This has been followed by a series of works that demonstrated the effectiveness of model-based approaches for solving video games~\citep{hafner2020mastering,ye2021mastering,kaiser2019model} and visual robot control tasks~\citep{ebert2018visual,hafner2019dream,seyde2020learning,rybkin2021model}.
There also have been several works that considered different objectives, including bisimulation~\citep{gelada2019deepmdp} and contrastive learning~\citep{nguyen2021temporal,okada2021dreaming,deng2021dreamerpro}. 
While most prior works optimize a single model to learn both visual representations and dynamics,
we instead develop a framework that decouples visual representation learning and dynamics learning. 

\paragraph{Self-supervised vision transformers}
Self-supervised learning with vision transformers (ViT)~\citep{dosovitskiy2020image} has been actively studied.
For instance,
\citet{chen2021empirical} introduced MoCo-v3 which trains a ViT with contrastive learning.
\citet{caron2021emerging} introduced DINO which utilizes a self-distillation loss~\citep{hinton2015distilling}, and demonstrated that self-supervised ViTs contain information about the semantic layout of images.
Training self-supervised ViTs with masked image modeling~\citep{he2021masked,bao2021beit,li2021mst,feichtenhofer2022masked,xie2022simmim,wei2022masked,zhou2021ibot} has also been successful.
In particular, \citet{he2021masked} proposed a masked autoencoder (MAE) that reconstructs masked pixel patches with an asymmetric encoder-decoder architecture.
Unlike MAE, we propose to randomly mask features from early convolutional layers~\citep{xiao2021early} instead of pixel patches and demonstrate that self-supervised ViTs can also be effective for visual model-based RL.

We provide more discussion on related works in more detail in~\cref{appendix:extended_related_work}.

\begin{figure*} [t!] \centering
    \hfill
    \subfigure[Pick Place]
    {
    \includegraphics[width=0.18\textwidth]{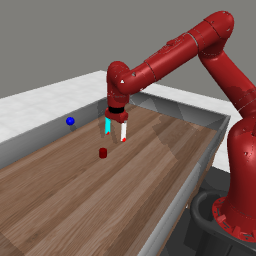}
    \label{fig:example_pick_place}}
    \hfill
    \subfigure[Shelf Place]
    {
    \includegraphics[width=0.18\textwidth]{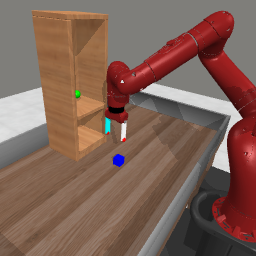}
    \label{fig:example_shelf_place}} 
    \hfill
    \subfigure[Reach Target]
    {
    \includegraphics[width=0.18\textwidth]{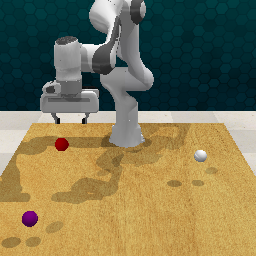}
    \label{fig:example_reach_target}} 
    \hfill
    \subfigure[Push Button]
    {
    \includegraphics[width=0.18\textwidth]{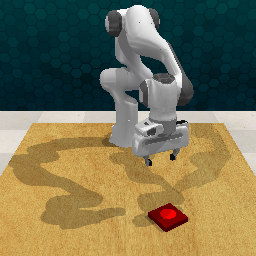}
    \label{fig:example_push_button}} 
    \hfill
    \subfigure[Reach Duplo]
    {
    \includegraphics[width=0.18\textwidth]{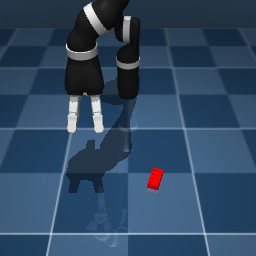}
    \label{fig:example_reach_duplo}}  
    \hfill
    \caption{
    Examples of visual observations used in our experiments. We consider a variety of visual robot control tasks from Meta-world~\citep{yu2020meta}, RLBench~\citep{james2020rlbench}, and DeepMind Control Suite~\citep{tassa2020dm_control}.}
    \label{fig:example}
    \vspace{-0.1in}
\end{figure*}

\section{Preliminaries}
\label{sec:preliminaries}
\paragraph{Problem formulation}
We formulate a visual control task as a partially observable Markov decision process (POMDP)~\citep{sutton2018reinforcement}, which is defined as a tuple $\left( \mathcal{O}, \mathcal{A}, p, r, \gamma\right)$.
$\mathcal{O}$ is the observation space, $\mathcal{A}$ is the action space, $p\left(o_{t}|o_{< t}, a_{< t}\right)$ is the transition dynamics, 
$r$ is the reward function that maps previous observations and actions to a reward $r_{t} = r\left(o_{\leq t}, a_{< t}\right)$,
and $\gamma \in [0,1)$ is the discount factor.

\paragraph{Dreamer} Dreamer \citep{hafner2020mastering,hafner2019dream} is a visual model-based RL method that learns world models from pixels and trains an actor-critic model via latent imagination. Specifically, Dreamer learns a Recurrent State Space Model (RSSM)~\citep{hafner2019learning}, which consists of following four components:
\begin{gather}
\begin{aligned}
&\text{Representation model:} &&s_t\sim q_\theta(s_{t} \,|\,s_{t-1},a_{t-1}, o_{t}) &&\text{Image decoder:} &&\hat{o}_t\sim p_\theta(\hat{o}_{t} \,|\,s_{t}) \\
&\text{Transition model:} &&\hat{s}_t\sim p_\theta(\hat{s}_{t} \,|\,s_{t-1}, a_{t-1}) &&\text{Reward predictor:} &&\hat{r}_t\sim p_\theta(\hat{r}_{t} \,|\,s_{t})
\label{eq:dreamer_world_model}
\end{aligned}
\end{gather}
The representation model extracts model state $s_{t}$ from previous model state $s_{t-1}$, previous action $a_{t-1}$, and current observation $o_{t}$. The transition model predicts future state $\hat{s}_{t}$ without the access to current observation $o_{t}$.
The image decoder reconstructs raw pixels to provide learning signal, and the reward predictor enables us to compute rewards from future model states without decoding future frames.
All model parameters $\theta$ are trained to jointly learn visual representations and environment dynamics by minimizing the negative variational lower bound~\citep{kingma2013auto}:
\begin{gather}
\begin{aligned}
    &\mathcal{L}(\theta) \doteq  \mathbb{E}_{q_{\theta}\left(s_{1:T}|a_{1:T},o_{1:T}\right)}\Big[ \\
    &\textstyle\sum_{t=1}^{T} \Big(
    -\ln p_{\theta}(o_{t}|s_{t})
    -\ln p_{\theta}(r_{t}|s_{t}) + \beta\,\text{KL}\left[ q_{\theta}(s_{t}|s_{t-1},a_{t-1},o_{t}) \,\Vert\,  p_{\theta}(\hat{s}_{t}|s_{t-1},a_{t-1}) \right]
    \Big)\Big],
    \label{eq:dreamer_objective}
\end{aligned}
\end{gather}
where $\beta$ is a hyperparameter that controls the tradeoff between the quality of visual representation learning and the accuracy of dynamics learning \citep{alemi2018fixing}.
Then, the critic is learned to regress the values computed from imaginary rollouts, and the actor is trained to maximize the values by propagating analytic gradients back through the transition model (see \cref{appendix:behavior_learning} for the details).

\paragraph{Masked autoencoder} Masked autoencoder (MAE)~\citep{he2021masked} is a self-supervised visual representation technique that trains an autoencoder to reconstruct raw pixels with randomly masked patches consisting of pixels.
Following a scheme introduced in vision transformer (ViT)~\citep{dosovitskiy2020image}, the observation $o_{t} \in \mathbb{R}^{H \times W \times C}$ is processed with a patchify stem that reshapes $o_{t}$ into a sequence of 2D patches $h_{t} \in \mathbb{R}^{N \times (P^{2} C)}$, where $P$ is the patch size and $N = HW/P^{2}$ is the number of patches.
Then a subset of patches is randomly masked with a ratio of $m$ to construct $h^{m}_{t} \in \mathbb{R}^{M \times (P^{2} C)}$.
\begin{gather}
\begin{aligned}
&\text{Patchify stem:} &&h_{t}= f^{\texttt{patch}}_{\phi}(o_{t}) &&\text{Masking:} &&h_{t}^{m}\sim p^{\texttt{mask}}(h_{t}^{m}\,|\,h_{t}, m)
\label{eq:mae_patchify}
\end{aligned}
\end{gather}
A ViT encoder embeds only the remaining patches $h^{m}_{t}$ into $D$-dimensional vectors,
concatenates the embedded tokens with a learnable CLS token, and processes them through a series of Transformer layers~\citep{vaswani2017attention}.
Finally, a ViT decoder reconstructs the observation by processing tokens from the encoder and learnable mask tokens through Transformer layers followed by a linear output head:
\begin{gather}
\begin{aligned}
&\text{ViT encoder:} &&z_{t}^{m}\sim p_{\phi}(z_{t}^{m} \,|\,h_{t}^{m})
&&\text{ViT decoder:} &&\hat{o}_t\sim p_\phi(\hat{o}_{t} \,|\,z_{t}^{m})
\label{eq:mae_vit}
\end{aligned}
\end{gather}
All the components paramaterized by $\phi$ are jointly optimized to minimize the mean squared error (MSE) between the reconstructed and original pixel patches.
MAE computes $z^{0}_{t}$ without masking, and utilizes its first component (\textit{i.e.,} CLS representation) for downstream tasks (\textit{e.g.,} image classification).

\section{Masked World Models}
\label{sec:method}
In this section, we present Masked World Models (\ALG), a visual model-based RL framework for learning accurate world models by separately learning visual representations and environment dynamics.
Our method repeats (i) updating an autoencoder with convolutional feature masking and an auxiliary reward prediction task (see~\cref{subsec:method_visual}), (ii) learning a dynamics model in the latent space of the autoencoder (see~\cref{subsec:method_dynamics}),
and (iii) collecting samples from environment interaction.
We provide the overview and pseudocode of \ALG in~\cref{fig:method_overview} and~\cref{appendix:pseudocode}, respectively.

\subsection{Visual Representation Learning}
\label{subsec:method_visual}
It has been observed that masked image modeling with a ViT architecture~\citep{he2021masked,bao2021beit,feichtenhofer2022masked} enables compute-efficient and stable self-supervised visual representation learning.
This motivates us to adopt this approach for visual model-based RL, but we find that masked image modeling with commonly used pixel patch masking~\citep{he2021masked} often makes it difficult to learn fine-grained details within patches, \textit{e.g.,} small objects (see~\cref{appendix:qualitative_analysis} for a motivating example).
While one can consider small-size patches, this would increase computational costs due to the quadratic complexity of self-attention layers.

To handle this issue, we instead propose to train an autoencoder that reconstructs raw pixels given randomly masked convolutional features.
Unlike previous approaches that utilize a patchify stem and randomly mask pixel patches (see~\cref{sec:preliminaries}),
we adopt a convolution stem~\citep{dosovitskiy2020image,xiao2021early} that processes $o_{t}$ through a series of convolutional layers followed by a flatten layer, to obtain $h^{c}_{t} \in \mathbb{R}^{N_{c} \times D}$ where $N_{c}$ is the number of convolutional features.
Then $h^{c}_{t}$ is randomly masked with a ratio of $m$ to obtain $h^{c,m}_{t} \in \mathbb{R}^{M_{c} \times D}$, and ViT encoder and decoder process $h^{c,m}_{t}$ to reconstruct raw pixels.
\begin{gather}
\begin{aligned}
&\text{Convolution stem:} &&h^{c}_{t}= f^{\texttt{conv}}_{\phi}(o_{t})
&&\text{Masking:} &&h_{t}^{c,m}\sim p^{\texttt{mask}}(h_{t}^{c,m}\,|\,h^{c}_{t}, m)
\\
&\text{ViT encoder:} &&z_{t}^{c,m}\sim p_{\phi}(z_{t}^{c,m} \,|\,h_{t}^{c,m})
&&\text{ViT decoder:} &&\hat{o}_t\sim p_\phi(\hat{o}_{t} \,|\,z_{t}^{c,m})
\label{eq:ours_conv}
\end{aligned}
\end{gather}
Because early convolutional layers mix low-level details, we find that our autoencoder can effectively reconstruct all the details within patches by learning to extract information from nearby non-masked features (see~\cref{fig:qualitative_analysis_prediction} for examples).
This enables us to learn visual representations capturing such details while also achieving the benefits of MAE, \textit{e.g.,} stability and compute-efficiency\footnote{We also note that \citet{xiao2021early} showed that introducing early convolutional layers can stabilize the training of ViT models, which can be helpful for RL that requires stability from the beginning of the training.}.

\paragraph{Reward prediction} In order to encode task-relevant information that might not be captured solely by the reconstruction objective,
we introduce an auxiliary objective for the autoencoder to predict rewards jointly with pixels.
Specifically, we make the autoencoder predict the reward $r_{t}$ from $z_{t}^{c,m}$ in conjunction with raw pixels.
\begin{gather}
\begin{aligned}
&\text{ViT decoder with reward prediction:} &&\hat{o}_t, \hat{r}_t\sim p_\phi(\hat{o}_{t}, \hat{r}_{t} \,|\,z_{t}^{c,m})
\label{eq:ours_reward}
\end{aligned}
\end{gather}
In practice, we concatenate one additional learnable mask token to inputs of the ViT decoder, and utilize the corresponding output representation for predicting the reward with a linear output head.

\paragraph{High masking ratio}
Introducing early convolutional layers might impede the masked reconstruction tasks because they propagate information across patches~\cite{touvron2022three}, and the model can exploit this to find a shortcut to solve reconstruction tasks.
However, we find that a high masking ratio (\textit{i.e.,} 75\%) can prevent the model from finding such shortcuts and induce useful representations (see~\cref{fig:metaworld_ablation_masking_ratio} for supporting experimental results).
This also aligns with the observation from~\citet{touvron2022three}, where masked image modeling~\citep{bao2021beit} with a convolution stem~\citep{graham2021levit} can achieve competitive performance with the patchify stem on the ImageNet classification task~\citep{deng2009imagenet}.

\subsection{Latent Dynamics Learning}
\label{subsec:method_dynamics}
Once we learn visual representations, we leverage them for efficiently learning a dynamics model in the latent space of the autoencoder.
Specifically, we obtain the frozen representations $z^{c,0}_{t}$ from the autoencoder, and then train a variant of RSSM whose inputs and reconstruction targets are $z^{c}_{t,0}$, by replacing the representation model and the image decoder in~\cref{eq:dreamer_world_model} with following components:
\begin{gather}
\begin{aligned}
&\text{Representation model:} &&s_t\sim q_\theta(s_{t} \,|\,s_{t-1},a_{t-1}, z^{c,0}_{t})\\
&\text{Visual representation decoder:} &&\hat{z}^{c,0}_{t}\sim p_\theta(\hat{z}^{c,0}_{t} \,|\,s_{t})
\label{eq:ours_world_model}
\end{aligned}
\end{gather}
Because visual representations capture both high- and low-level information in an abstract form, the model can focus more on dynamics learning by reconstructing them instead of raw pixels (see \cref{subsec:experiments_qualitative_analysis} for relevant discussion).
Here, we also note that we utilize all the elements of $z_{t}^{c,0}$ unlike MAE that only utilizes CLS representation for downstream tasks.
We empirically find this enables the model to receive rich learning signals from reconstructing all the representations containing spatial information (see~\cref{appendix:full_ablation_analysis} for supporting experiments).

\paragraph{Optimization}
Given a random batch $\{(o_{j}, r_{j}, a_{j})\}_{j=1}^{B}$, MWM objective is defined as follows:
\vspace{-0.1in}
\begin{align}
    &\mathcal{L}^{\texttt{mwm}}(\phi,\theta) = \frac{1}{B}\sum_{j=1}^{B}\Big(\describe{-\ln p_{\phi}(o_{j}|z^{c,m}_{j}) -\ln p_{\phi}(r_{j}|z^{c,m}_{j})}{visual representation learning}\\
    &\quad\quad\describe{-\ln p_{\theta}(z^{c,0}_{j}|s_{j})-\ln p_{\theta}(r_{j}|s_{j}) +  \beta\,\text{KL}\left[ q_{\theta}(s_{j}|s_{j-1},a_{j-1},z^{c,0}_{j}) \,\Vert\,  p_{\theta}(\hat{s}_{j}|s_{j-1},a_{j-1}) \right]}{dynamics learning}\Big)\nonumber
\end{align}

\section{Experiments}
\label{sec:experiments}
We evaluate \ALG on various robotics benchmarks, including Meta-world~\citep{yu2020meta} (see~\cref{subsec:experiments_meta_world}), RLBench~\citep{james2020rlbench} (see~\cref{subsec:experiment_rlbench}), and DeepMind Control Suite~\citep{tassa2012synthesis} (see~\cref{subsec:experiments_dmc}).
We remark that these benchmarks consist of diverse and challenging visual robotic tasks.
We also analyze algorithmic design choices in-depth (see~\cref{subsec:experiments_ablation}) and provide a qualitative analysis of how our decoupling approach works by visualizing the predictions from the latent dynamics model (see~\cref{subsec:experiments_qualitative_analysis}).

\begin{figure*} [t!] \centering
    \includegraphics[width=0.99\textwidth]{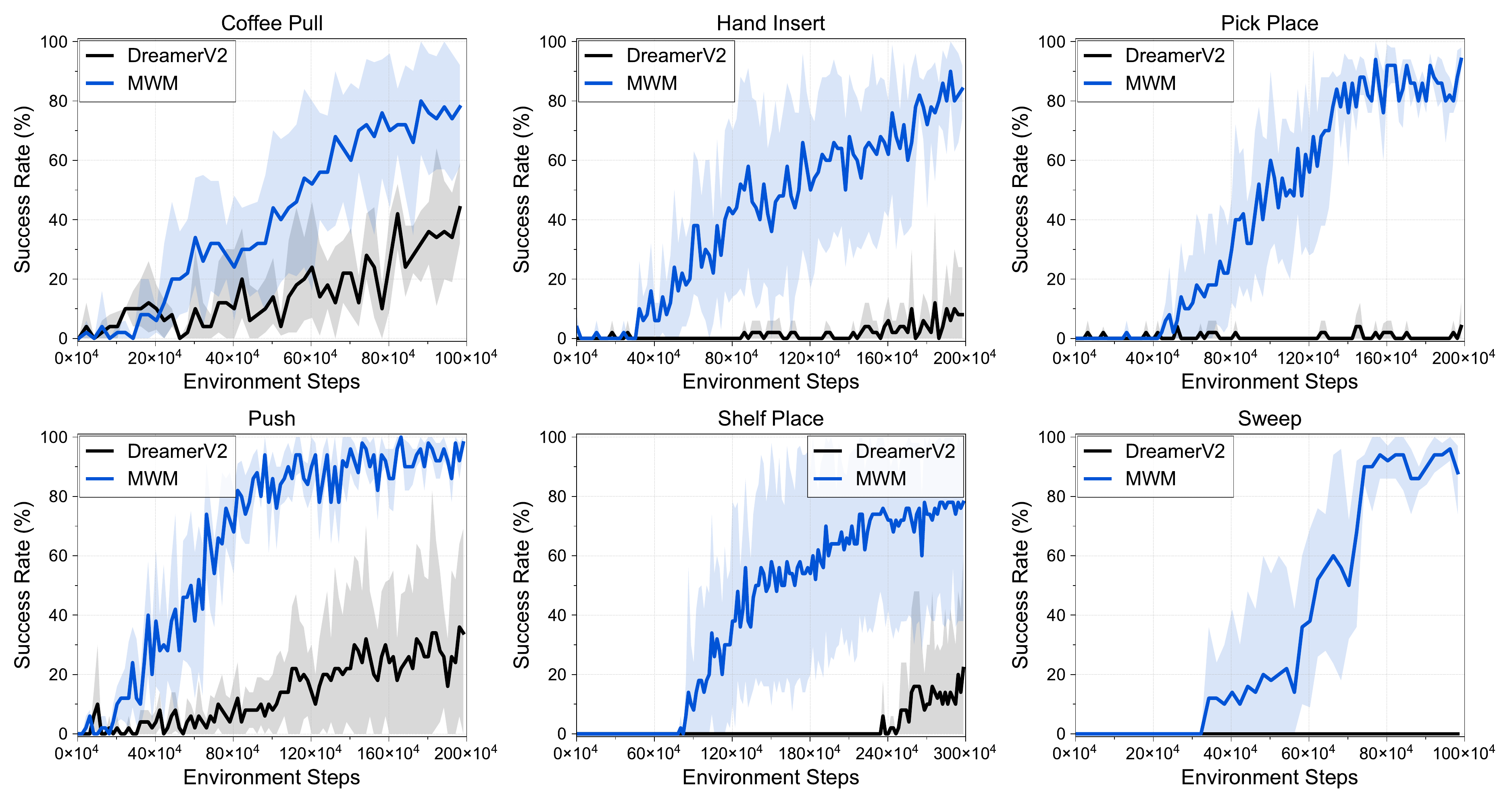}
    \vspace{-0.1in}
    \caption{
    Learning curves on six visual robotic manipulation tasks from Meta-world as measured on the success rate.
    We select the tasks that require modeling interactions between small objects and robot arms.
    Learning curves on 50 tasks are available in~\cref{appendix:full_meta_world}.
    The solid line and shaded regions represent the mean and bootstrap confidence intervals, respectively, across five runs.}
    \label{fig:metaworld_difficult}
    \vspace{-0.05in}
\end{figure*}

\paragraph{Implementation}
We use visual observations of $64 \times 64 \times 3$.
For the convolution stem,
we stack 3 convolution layers with the kernel size of 4 and stride 2, followed by a linear projection layer.
We use a 4-layer ViT encoder and a 3-layer ViT decoder.
We find that initializing the autoencoder with a warm-up schedule at the beginning of training is helpful.
Unlike MAE, we compute the loss on entire pixels because we do not apply masking to pixels.
For world models, we build our implementation on top of DreamerV2~\citep{hafner2020mastering}.
To take a sequence of autoencoder representations as inputs, we replace a CNN encoder and decoder with a 2-layer Transformer encoder and decoder.
We use same hyperparameters within the same benchmark.
More details are available in~\cref{appendix:architecture_details}.

\subsection{Meta-world Experiments}
\label{subsec:experiments_meta_world}
\paragraph{Environment details}
In order to use a single camera viewpoint consistently over all 50 tasks, we use the modified $\texttt{corner2}$ camera viewpoint for all tasks.
In our experiments, we classify 50 tasks into $\texttt{easy}$, $\texttt{medium}$, $\texttt{hard}$, and $\texttt{very hard}$ tasks where experiments are run over 500K, 1M, 2M, 3M environments steps with action repeat of 2, respectively.
More details are available in~\cref{appendix:experiments_details}.

\begin{figure*} [t!] \centering
    \subfigure[Meta-world aggregated]
    {
    \includegraphics[width=0.315\textwidth]{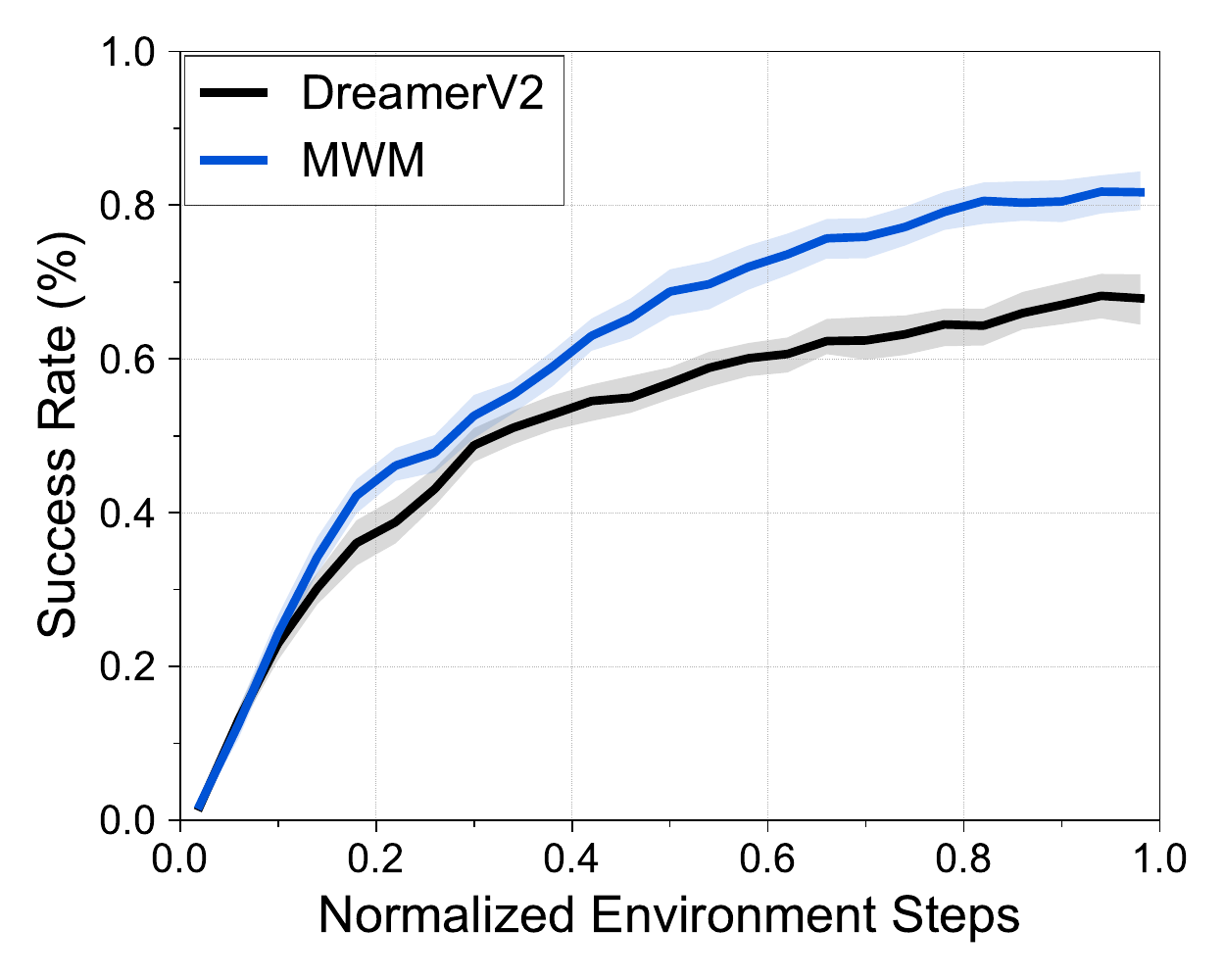}
    \label{fig:metaworld_aggregate}}
    \subfigure[RLBench: Reach Target]
    {
    \includegraphics[width=0.315\textwidth]{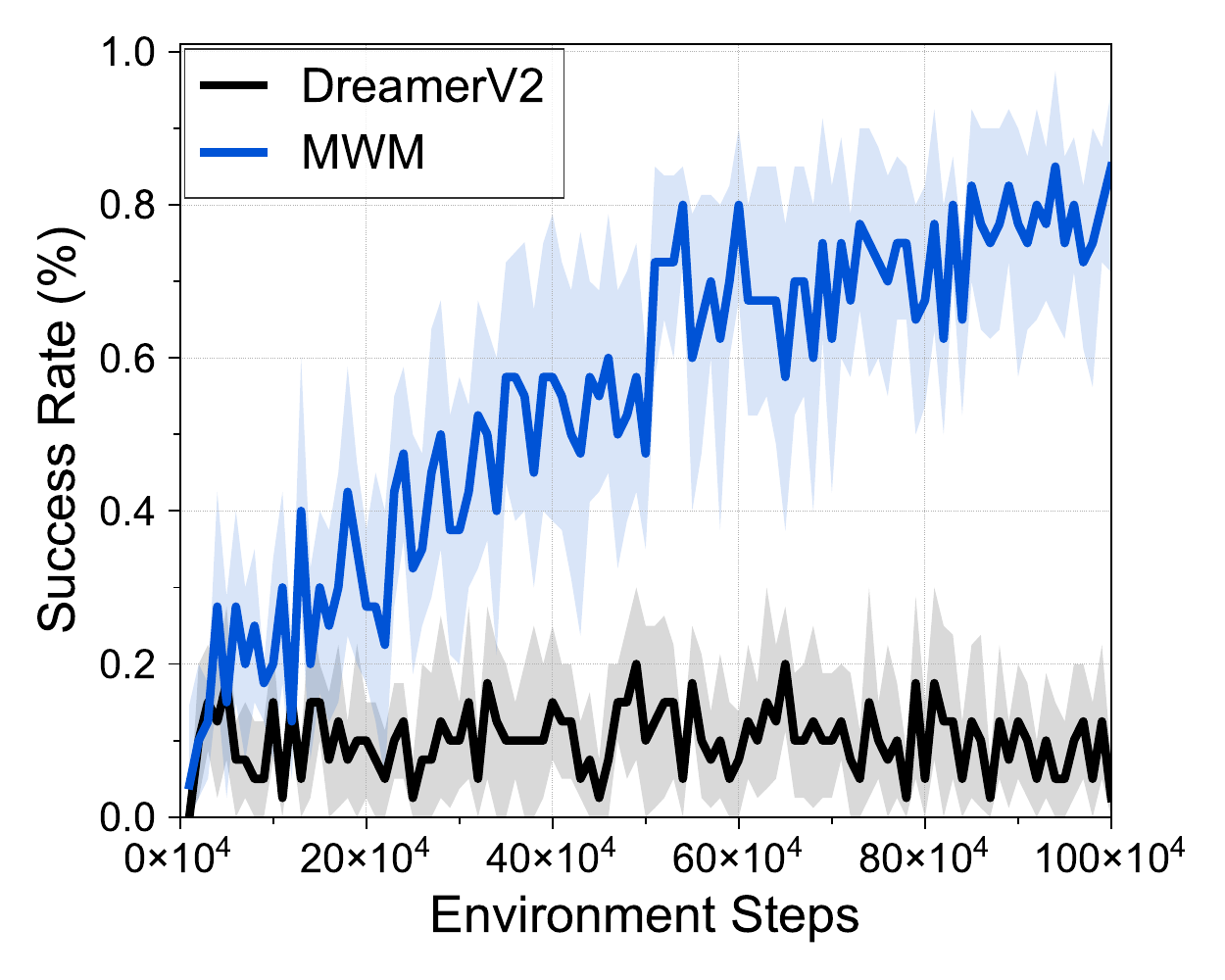}
    \label{fig:rlbench_reach_target}}
    \subfigure[RLBench: Push Button]
    {
    \includegraphics[width=0.315\textwidth]{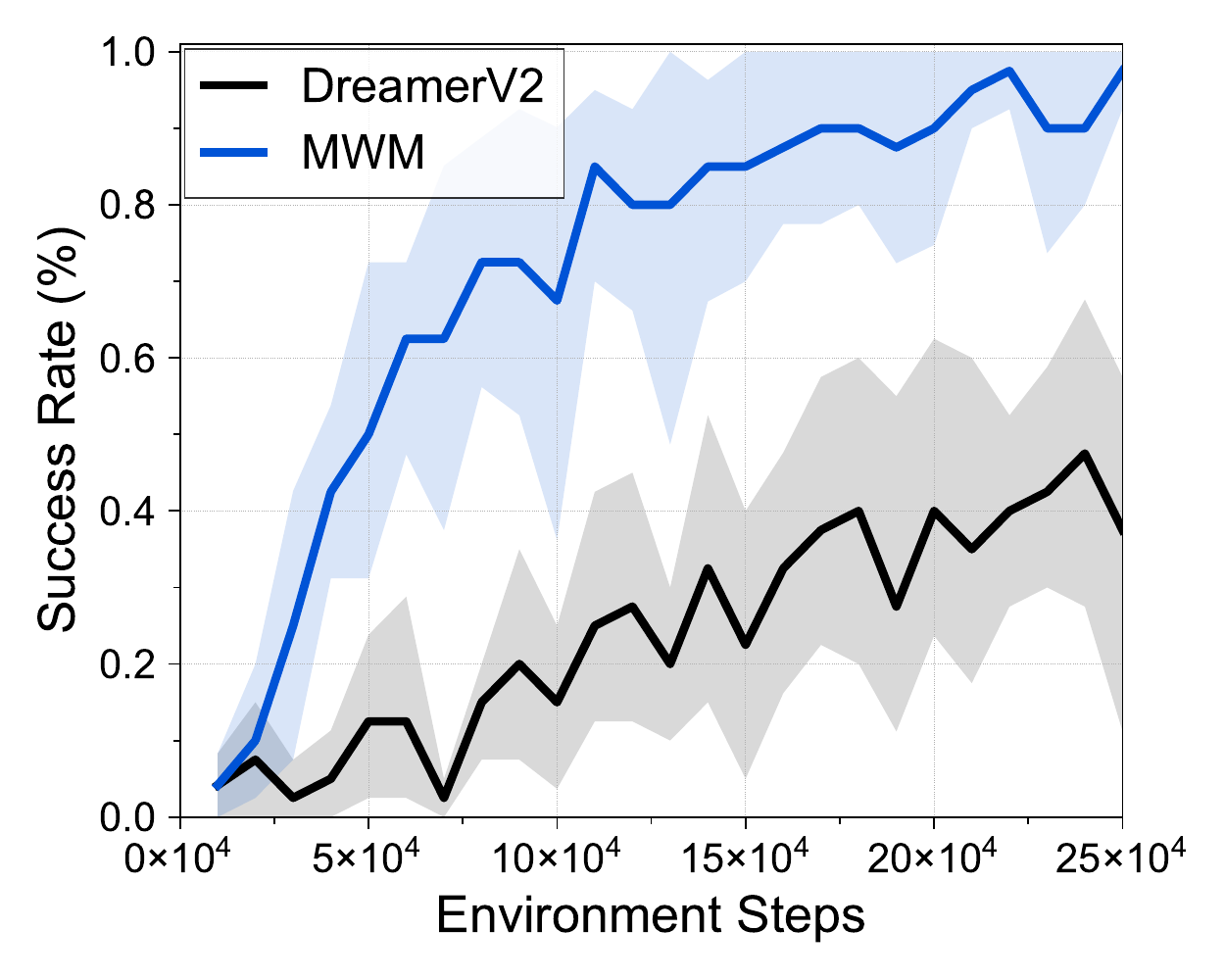}
    \label{fig:rlbench_push_button}}
    \vspace{-0.05in}
    \caption{
    (a) Aggregate performance on all 50 Meta-world tasks. We normalize environment steps by maximum steps in each task. The solid line and shaded regions represent the mean and stratified bootstrap confidence intervals, respectively, across 250 runs.
    We report the learning curves on (b) Reach Target and (c) Push Button from RLBench.
    Performances are not directly comparable to previous results~\citep{james2022q,james2021coarse} due to the difference in setups (see~\cref{subsec:experiment_rlbench}).
    The solid line and shaded regions represent the mean and bootstrap confidence intervals, respectively, across eight runs.
    }
    \label{fig:figure_collection}
    \vspace{-0.05in}
\end{figure*}

\paragraph{Results}
In~\cref{fig:metaworld_difficult}, we report the performance on a set of selected six challenging tasks that require agents to control robot arms to interact with small objects.
We find that \ALG significantly outperforms DreamerV2 in terms of both sample-efficiency and final performance.
In particular, \ALG achieves $>80\%$ success rate on Pick Place while DreamerV2 struggles to solve the task.
These results show that our approach of separating visual representation learning and dynamics learning can learn accurate world models on challenging domains.
We also note that DreamerV2 learns visual representations with the reward prediction loss, which implies that our superior performance does not come from utilizing additional information.
\cref{fig:metaworld_aggregate} shows the aggregate performance over all the 50 tasks from the benchmark, demonstrating that our method consistently outperforms DreamerV2 overall.
We also provide learning curves on all individual tasks in~\cref{appendix:full_meta_world}, where \ALG consistently achieves similar or better performance on most tasks.

\subsection{RLBench Experiments}
\label{subsec:experiment_rlbench}
\paragraph{Environment details}
In order to evaluate our method on more challenging visual robotic manipulation tasks, we consider RLBench~\citep{james2020rlbench}, which has previously acted as an effective proxy for real-robot performance~\citep{james2021coarse}.
Since RLBench consists of sparse-reward and challenging tasks, solving them typically requires expert demonstrations, specialized network architectures, additional inputs (e.g., point cloud and proprioceptive states), and an action mode that requires path planning~\citep{james2022q,james2021coarse,james2022lpr,james2022tree}.
While we could utilize some of these components, we instead leave this as future work in order to maintain a consistent evaluation setup across multiple domains.
In our experiments, we instead consider two relatively easy tasks with dense rewards, and utilize an action mode that specifies the delta of joint positions. We provide more details in~\cref{appendix:experiments_details}.

\paragraph{Results}
As shown in~\cref{fig:rlbench_reach_target} and~\cref{fig:rlbench_push_button}, we observe that our approach can also be effective on RLBench tasks, significantly outperforming DreamerV2.
In particular, DreamerV2 achieves $<20\%$ success rate on Reach Target, while our approach can solve the tasks with $>80\%$ success rates.
We find that this is because DreamerV2 fails to capture target positions in visual observations, while our method can capture such details (see~\cref{subsec:experiments_qualitative_analysis} for relevant discussion and visualizations).
However, we also note that these results are preliminary because they are still too sample-inefficient to be used for real-world scenarios.
We provide more discussion in~\cref{sec:discussion}.

\begin{figure*} [t!] \centering
    \includegraphics[width=0.99\textwidth]{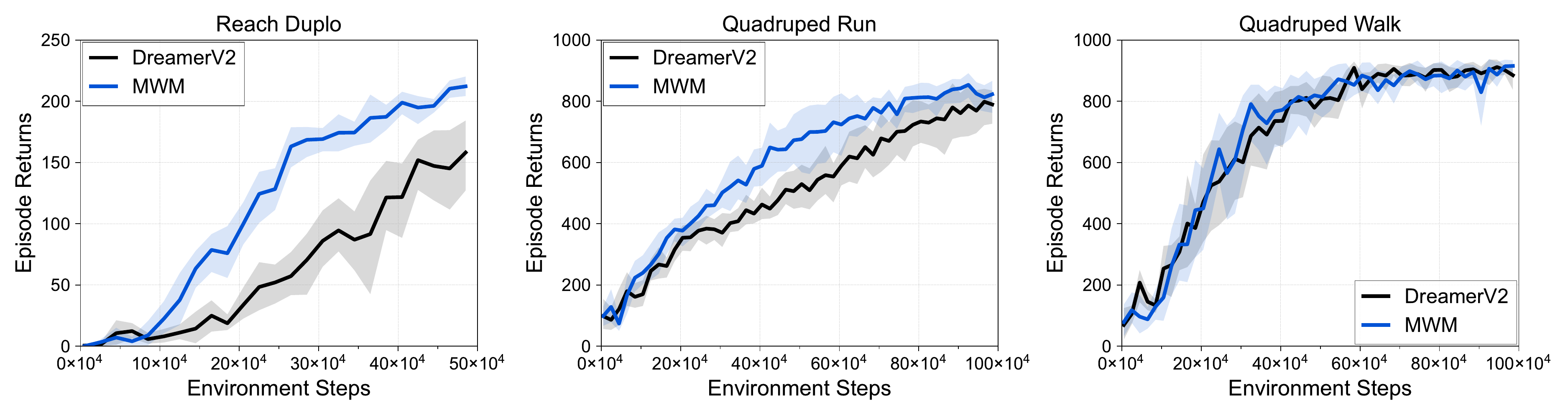}
    \vspace{-0.05in}
    \caption{
    Learning curves on three visual robot control tasks from DeepMind Control Suite as measured on the episode return.
    The solid line and shaded regions represent the mean and bootstrap confidence intervals, respectively, across eight runs.}
    \label{fig:dmc}
    % \vspace{-0.2125in}
    \vspace{-0.15in}
\end{figure*}

\subsection{DeepMind Control Suite Experiments}
\label{subsec:experiments_dmc}
\paragraph{Environment details} In order to demonstrate that our approach is generally applicable to diverse visual control tasks,
we also evaluate our method on visual locomotion tasks from the widely used DeepMind Control Suite benchmark.
Following a standard setup in~\citet{hafner2019dream}, we use an action repeat of 2 and default camera configurations.
We provide more details in~\cref{appendix:experiments_details}.

\paragraph{Results}
\cref{fig:dmc} shows that our method achieves competitive performance to DreamerV2 on visual locomotion tasks (i.e., Quadruped tasks), demonstrating the generality of our approach across diverse visual control tasks.
We also observe that our method outperforms DreamerV2 on Reach Duplo, which is one of a few manipulation tasks in the benchmark (see~\cref{fig:example_reach_duplo} for an example).
This implies that our method is effective on environments where the model should capture fine-grained details like object positions.
More results are available in~\cref{appendix:additional_dmc}, where trends are similar.

\subsection{Ablation Study}
\label{subsec:experiments_ablation}
\begin{figure*} [t!] \centering
    \subfigure[Feature masking]
    {
    \includegraphics[width=0.315\textwidth]{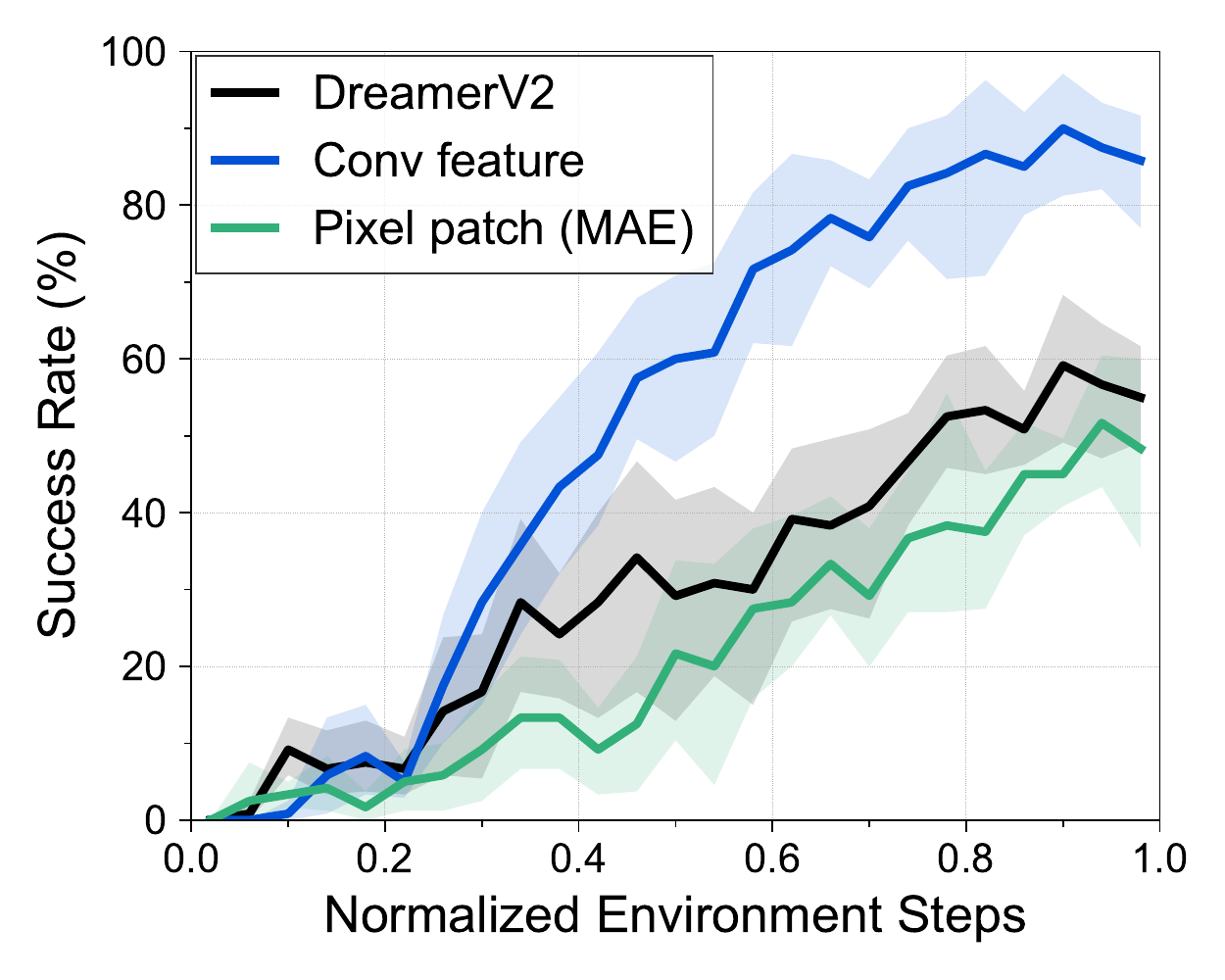}
    \label{fig:metaworld_ablation_mae}}
        \subfigure[Masking ratio]
    {
    \includegraphics[width=0.315\textwidth]{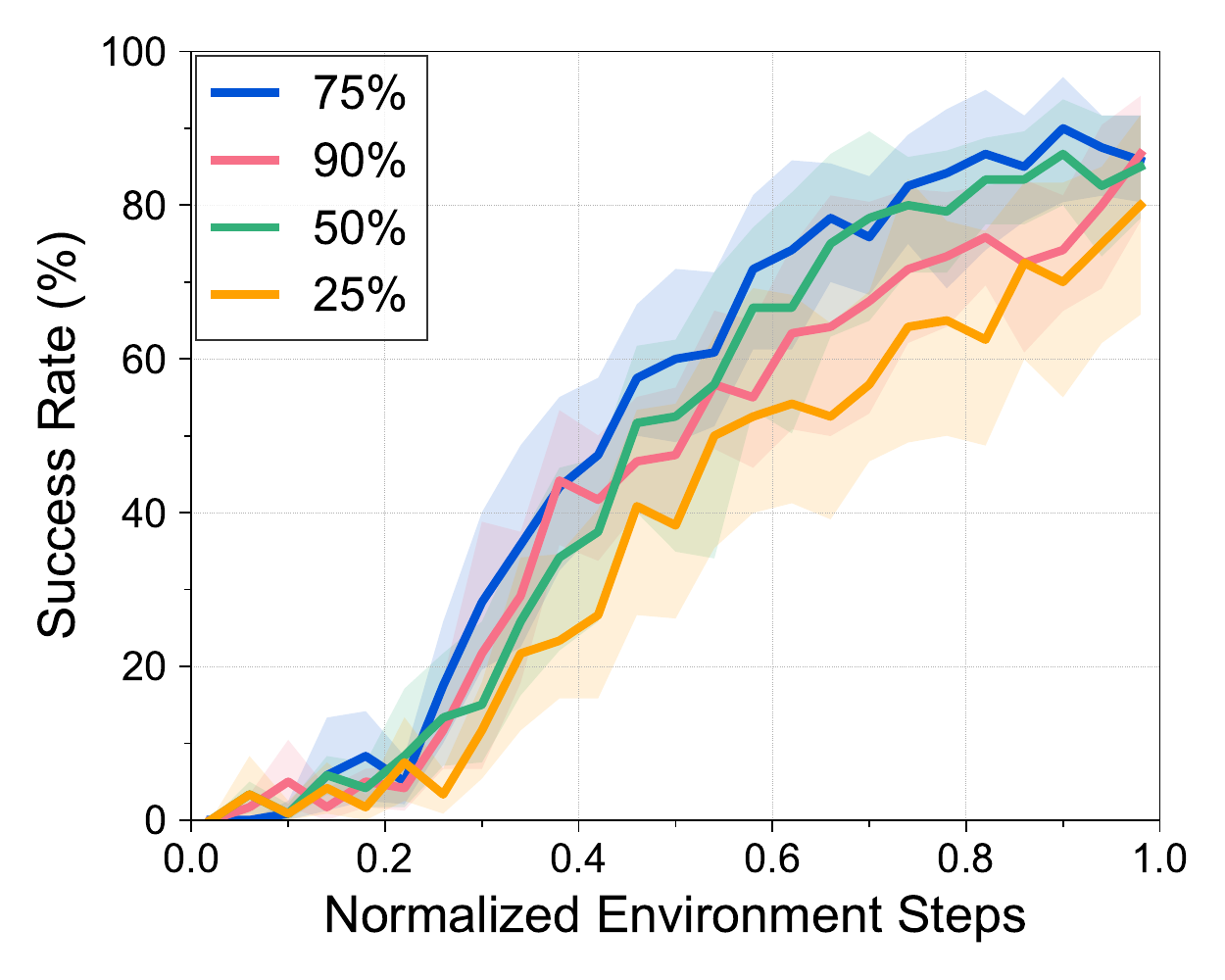}
    \label{fig:metaworld_ablation_masking_ratio}}
    \subfigure[Reward prediction]
    {
    \includegraphics[width=0.315\textwidth]{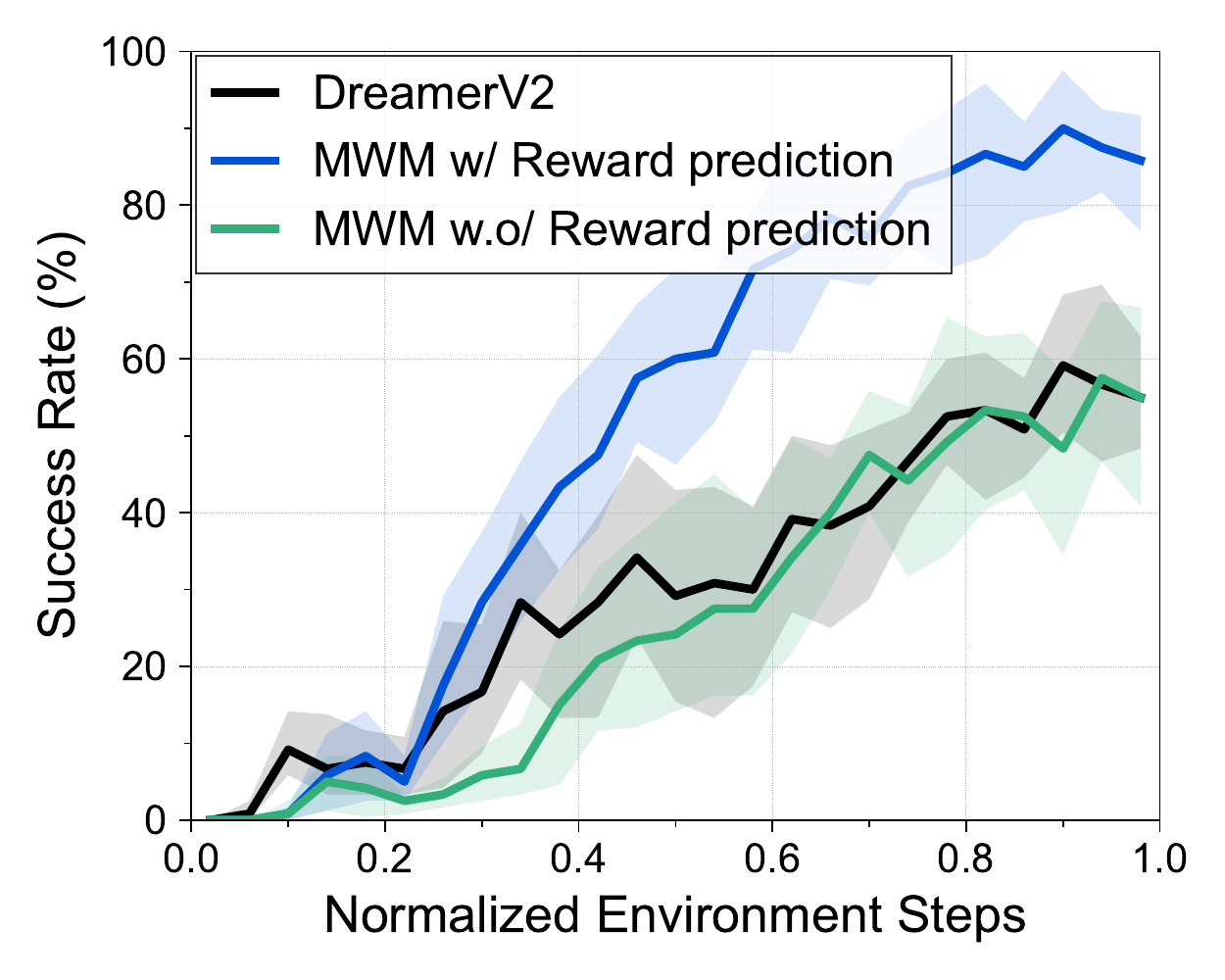}
    \label{fig:metaworld_ablation_reward}}
    \vspace{-0.05in}
    \caption{
    Learning curves on three manipulation tasks from Meta-world that investigate the effect of (a) convolutional feature masking, (b) masking ratio, and (c) reward prediction.
    The solid line and shaded regions represent the mean and stratified bootstrap confidence interval across 12 runs.
    }
    \label{fig:metaworld_ablation}
\end{figure*}

\paragraph{Convolutional feature masking}
We compare convolutional feature masking with pixel masking (\textit{i.e.,} MAE) in~\cref{fig:metaworld_ablation_mae}, which shows that convolutional feature masking significantly outperforms pixel masking.
Both results are achieved with the reward prediction.
This demonstrates that enabling the model to capture fine-grained details within patches can be important for visual control.
We also report the performance with varying masking ratio $m \in \{0.25, 0.5, 0.75, 0.9\}$ in~\cref{fig:metaworld_ablation_masking_ratio}.
As we discussed in~\cref{subsec:method_visual}, we find that $m=0.75$ achieves better performance than $m \in \{0.25, 0.5\}$ because strong regularization can prevent the model from finding a shortcut from input pixels.
However, we also find that too strong regularization (\textit{i.e.,} $m=0.9$) degrades the performance.

\begin{figure*} [t!] \centering
    \includegraphics[width=0.99\textwidth]{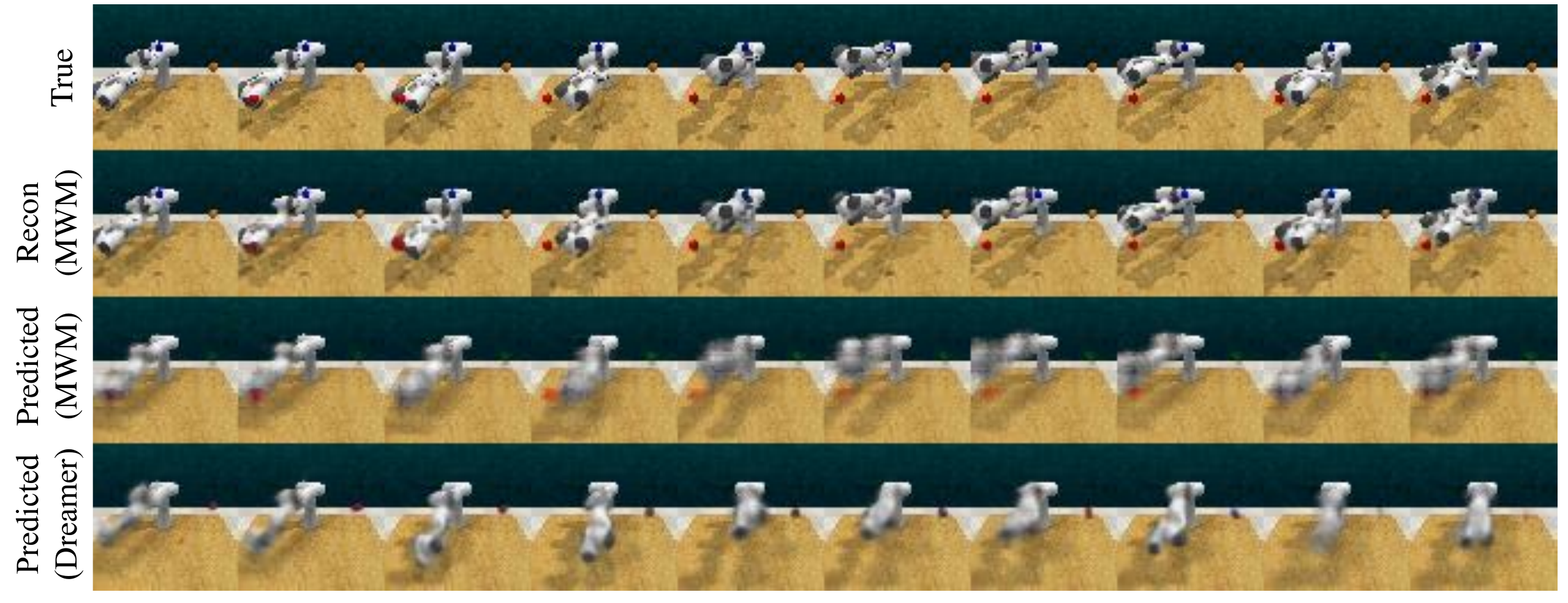}
    \caption{
    Future frames reconstructed with the autoencoder (\textit{i.e.,} Recon) and predicted by latent dynamics models (\textit{i.e.,} Predicted).
    Best viewed as video provided in~\cref{appendix:qualitative_analysis}.
    }
    \label{fig:qualitative_analysis_prediction}
    \vspace{-0.1in}
\end{figure*}

\paragraph{Reward prediction}
In~\cref{fig:metaworld_ablation_reward}, we find that performance significantly degrades without reward prediction, which shows that the reconstruction objective might not be sufficient for learning task-relevant information.
It would be an interesting future direction to develop a representation learning scheme that learns task-relevant information without rewards because they might not be available in practice.
We provide more ablation studies and learning curves on individual tasks in~\cref{appendix:full_ablation_analysis}. Moreover, we also show that reward prediction allows for learning useful representations that can be generally applicable to diverse robotic manipulation tasks in~\cref{appendix:generalizability}.

\subsection{Qualitative Analysis}
\label{subsec:experiments_qualitative_analysis}
We visually investigate how our world model works compared to the world model of DreamerV2.
Specifically, we visualize the future frames predicted by latent dynamics models on Reach Target from RLBench in~\cref{fig:qualitative_analysis_prediction}.
In this task, a robot arm should reach a target position specified by a red block in visual observations (see~\cref{fig:example_reach_target}), which changes every trial.
Thus it is crucial for the model to accurately predict the position of red blocks for solving the tasks.
We find that our world model effectively captures the position of red blocks, while DreamerV2 fails.
Interestingly, we also observe that our latent dynamics model ignores the components that are not task-relevant such as blue and orange blocks, though the reconstructions from the autoencoder are capturing all the details.
This shows how our decoupling approach works: it encourages the autoencoder to focus on learning representations capturing the details and the dynamics model to focus on modeling task-relevant components of environments.
We provide more examples in~\cref{appendix:qualitative_analysis}.

\section{Discussion}
\label{sec:discussion}
We have presented Masked World Models (\ALG), which is a visual model-based RL framework that decouples visual representation learning and dynamics learning.
By learning a latent dynamics model operating in the latent space of a self-supervised ViT, we find that our approach allows for solving a variety of visual control tasks from Meta-world, RLBench, and DeepMind Control Suite.

\paragraph{Limitation} Despite the results, there are a number of areas for improvement.
As we have shown in~\cref{fig:metaworld_ablation_reward}, the performance of our approach heavily depends on the auxiliary reward prediction task.
This might be because our autoencoder is not learning temporal information, which is crucial for learning task-relevant information.
It would be interesting to investigate the performance of video representation learning with ViTs~\citep{feichtenhofer2022masked,tong2022videomae,arnab2021vivit}.
It would also be interesting to study introducing auxiliary prediction for other modalities, such as audio.
Another weakness is that our model operates only on RGB pixels from a single camera viewpoint; we look forward to a future work that incorporates different input modalities such as proprioceptive states and point clouds, building on top of the recent multi-modal learning approaches~\citep{geng2022multimodal,bachmann2022multimae}.
Finally, our approach trains behaviors from scratch, which makes it still too sample-inefficient to be used in real-world scenarios.
Leveraging a small number of demonstrations, incorporating the action mode with path planning~\citep{james2022q}, or pre-training a world model on video datasets~\citep{seo2022reinforcement} are directions we hope to investigate in future works.

\acknowledgments{We would like to thank Jongjin Park and Sihyun Yu for helpful discussions. We also thank Cirrascale Cloud Services\footnote{\url{https://cirrascale.com}} for providing compute resources. 
This work was partially supported by National Science Foundation under grant NSF NRI \#2024675, Office of Naval Research under grant N00014-21-1-2769 and grant N00014-22-1-2121, the Darpa RACER program, the Hong Kong Centre for Logistics Robotics, BMW, and Institute of Information \& Communications Technology Planning \& Evaluation (IITP) grant funded by the Korea government (MSIT) (No.2019-0-00075, Artificial Intelligence Graduate School Program (KAIST)).}

\bibliography{main}

\begin{thebibliography}{88}
\providecommand{\natexlab}[1]{#1}
\providecommand{\url}[1]{\texttt{#1}}
\expandafter\ifx\csname urlstyle\endcsname\relax
  \providecommand{\doi}[1]{doi: #1}\else
  \providecommand{\doi}{doi: \begingroup \urlstyle{rm}\Url}\fi

\bibitem[Chua et~al.(2018)Chua, Calandra, McAllister, and Levine]{chua2018deep}
K.~Chua, R.~Calandra, R.~McAllister, and S.~Levine.
\newblock Deep reinforcement learning in a handful of trials using
  probabilistic dynamics models.
\newblock In \emph{Advances in neural information processing systems}, 2018.

\bibitem[Deisenroth and Rasmussen(2011)]{deisenroth2011pilco}
M.~Deisenroth and C.~E. Rasmussen.
\newblock Pilco: A model-based and data-efficient approach to policy search.
\newblock In \emph{International Conference on Machine Learning}, 2011.

\bibitem[Lenz et~al.(2015)Lenz, Knepper, and Saxena]{lenz2015deepmpc}
I.~Lenz, R.~A. Knepper, and A.~Saxena.
\newblock Deepmpc: Learning deep latent features for model predictive control.
\newblock In \emph{Robotics: Science and Systems}, 2015.

\bibitem[Kurutach et~al.(2018)Kurutach, Clavera, Duan, Tamar, and
  Abbeel]{kurutach2018model}
T.~Kurutach, I.~Clavera, Y.~Duan, A.~Tamar, and P.~Abbeel.
\newblock Model-ensemble trust-region policy optimization.
\newblock In \emph{International Conference on Learning Representations}, 2018.

\bibitem[Janner et~al.(2019)Janner, Fu, Zhang, and Levine]{janner2019trust}
M.~Janner, J.~Fu, M.~Zhang, and S.~Levine.
\newblock When to trust your model: Model-based policy optimization.
\newblock \emph{Advances in Neural Information Processing Systems}, 2019.

\bibitem[Finn and Levine(2017)]{finn2017deep}
C.~Finn and S.~Levine.
\newblock Deep visual foresight for planning robot motion.
\newblock In \emph{2017 IEEE International Conference on Robotics and
  Automation (ICRA)}, 2017.

\bibitem[Ebert et~al.(2018)Ebert, Finn, Dasari, Xie, Lee, and
  Levine]{ebert2018visual}
F.~Ebert, C.~Finn, S.~Dasari, A.~Xie, A.~Lee, and S.~Levine.
\newblock Visual foresight: Model-based deep reinforcement learning for
  vision-based robotic control.
\newblock \emph{arXiv preprint arXiv:1812.00568}, 2018.

\bibitem[Watter et~al.(2015)Watter, Springenberg, Boedecker, and
  Riedmiller]{watter2015embed}
M.~Watter, J.~Springenberg, J.~Boedecker, and M.~Riedmiller.
\newblock Embed to control: A locally linear latent dynamics model for control
  from raw images.
\newblock In \emph{Advances in neural information processing systems}, 2015.

\bibitem[Hafner et~al.(2019)Hafner, Lillicrap, Fischer, Villegas, Ha, Lee, and
  Davidson]{hafner2019learning}
D.~Hafner, T.~Lillicrap, I.~Fischer, R.~Villegas, D.~Ha, H.~Lee, and
  J.~Davidson.
\newblock Learning latent dynamics for planning from pixels.
\newblock In \emph{International Conference on Machine Learning}, 2019.

\bibitem[Zhang et~al.(2019)Zhang, Vikram, Smith, Abbeel, Johnson, and
  Levine]{zhang2019solar}
M.~Zhang, S.~Vikram, L.~Smith, P.~Abbeel, M.~Johnson, and S.~Levine.
\newblock Solar: Deep structured representations for model-based reinforcement
  learning.
\newblock In \emph{International Conference on Machine Learning}, 2019.

\bibitem[Ha and Schmidhuber(2018)]{ha2018world}
D.~Ha and J.~Schmidhuber.
\newblock World models.
\newblock In \emph{Advances in Neural Information Processing Systems}, 2018.

\bibitem[Kingma and Welling(2014)]{kingma2013auto}
D.~P. Kingma and M.~Welling.
\newblock Auto-encoding variational bayes.
\newblock In \emph{International Conference on Learning Representations}, 2014.

\bibitem[He et~al.(2021)He, Chen, Xie, Li, Doll{\'a}r, and
  Girshick]{he2021masked}
K.~He, X.~Chen, S.~Xie, Y.~Li, P.~Doll{\'a}r, and R.~Girshick.
\newblock Masked autoencoders are scalable vision learners.
\newblock \emph{arXiv preprint arXiv:2111.06377}, 2021.

\bibitem[Dosovitskiy et~al.(2021)Dosovitskiy, Beyer, Kolesnikov, Weissenborn,
  Zhai, Unterthiner, Dehghani, Minderer, Heigold, Gelly,
  et~al.]{dosovitskiy2020image}
A.~Dosovitskiy, L.~Beyer, A.~Kolesnikov, D.~Weissenborn, X.~Zhai,
  T.~Unterthiner, M.~Dehghani, M.~Minderer, G.~Heigold, S.~Gelly, et~al.
\newblock An image is worth 16x16 words: Transformers for image recognition at
  scale.
\newblock In \emph{International Conference on Learning Representaitons}, 2021.

\bibitem[Hafner et~al.(2021)Hafner, Lillicrap, Norouzi, and
  Ba]{hafner2020mastering}
D.~Hafner, T.~Lillicrap, M.~Norouzi, and J.~Ba.
\newblock Mastering atari with discrete world models.
\newblock In \emph{International Conference on Learning Representations}, 2021.

\bibitem[Yu et~al.(2020)Yu, Quillen, He, Julian, Hausman, Finn, and
  Levine]{yu2020meta}
T.~Yu, D.~Quillen, Z.~He, R.~Julian, K.~Hausman, C.~Finn, and S.~Levine.
\newblock Meta-world: A benchmark and evaluation for multi-task and meta
  reinforcement learning.
\newblock In \emph{Conference on Robot Learning}, 2020.

\bibitem[James et~al.(2020)James, Ma, Arrojo, and Davison]{james2020rlbench}
S.~James, Z.~Ma, D.~R. Arrojo, and A.~J. Davison.
\newblock R{LB}ench: The robot learning benchmark \& learning environment.
\newblock \emph{IEEE Robotics and Automation Letters}, 2020.

\bibitem[Touvron et~al.(2022)Touvron, Cord, El-Nouby, Verbeek, and
  J{\'e}gou]{touvron2022three}
H.~Touvron, M.~Cord, A.~El-Nouby, J.~Verbeek, and H.~J{\'e}gou.
\newblock Three things everyone should know about vision transformers.
\newblock \emph{arXiv preprint arXiv:2203.09795}, 2022.

\bibitem[Deng et~al.(2009)Deng, Dong, Socher, Li, Li, and
  Fei-Fei]{deng2009imagenet}
J.~Deng, W.~Dong, R.~Socher, L.-J. Li, K.~Li, and L.~Fei-Fei.
\newblock Imagenet: A large-scale hierarchical image database.
\newblock In \emph{2009 IEEE conference on computer vision and pattern
  recognition}, 2009.

\bibitem[Finn et~al.(2016)Finn, Tan, Duan, Darrell, Levine, and
  Abbeel]{finn2016deep}
C.~Finn, X.~Y. Tan, Y.~Duan, T.~Darrell, S.~Levine, and P.~Abbeel.
\newblock Deep spatial autoencoders for visuomotor learning.
\newblock In \emph{2016 IEEE International Conference on Robotics and
  Automation (ICRA)}, 2016.

\bibitem[Hafner et~al.(2020)Hafner, Lillicrap, Ba, and
  Norouzi]{hafner2019dream}
D.~Hafner, T.~Lillicrap, J.~Ba, and M.~Norouzi.
\newblock Dream to control: Learning behaviors by latent imagination.
\newblock In \emph{International Conference on Learning Representations}, 2020.

\bibitem[Kaiser et~al.(2019)Kaiser, Babaeizadeh, Milos, Osinski, Campbell,
  Czechowski, Erhan, Finn, Kozakowski, Levine, et~al.]{kaiser2019model}
L.~Kaiser, M.~Babaeizadeh, P.~Milos, B.~Osinski, R.~H. Campbell, K.~Czechowski,
  D.~Erhan, C.~Finn, P.~Kozakowski, S.~Levine, et~al.
\newblock Model-based reinforcement learning for atari.
\newblock In \emph{International Conference on Learning Representations}, 2019.

\bibitem[Gupta et~al.(2022)Gupta, Tian, Zhang, Wu, Mart{\'\i}n-Mart{\'\i}n, and
  Fei-Fei]{gupta2022maskvit}
A.~Gupta, S.~Tian, Y.~Zhang, J.~Wu, R.~Mart{\'\i}n-Mart{\'\i}n, and L.~Fei-Fei.
\newblock Maskvit: Masked visual pre-training for video prediction.
\newblock \emph{arXiv preprint arXiv:2206.11894}, 2022.

\bibitem[Ye et~al.(2021)Ye, Liu, Kurutach, Abbeel, and Gao]{ye2021mastering}
W.~Ye, S.~Liu, T.~Kurutach, P.~Abbeel, and Y.~Gao.
\newblock Mastering atari games with limited data.
\newblock \emph{Advances in Neural Information Processing Systems}, 2021.

\bibitem[Seyde et~al.(2021)Seyde, Schwarting, Karaman, and
  Rus]{seyde2020learning}
T.~Seyde, W.~Schwarting, S.~Karaman, and D.~Rus.
\newblock Learning to plan optimistically: Uncertainty-guided deep exploration
  via latent model ensembles.
\newblock In \emph{Conference on Robot Learning}, 2021.

\bibitem[Rybkin et~al.(2021)Rybkin, Zhu, Nagabandi, Daniilidis, Mordatch, and
  Levine]{rybkin2021model}
O.~Rybkin, C.~Zhu, A.~Nagabandi, K.~Daniilidis, I.~Mordatch, and S.~Levine.
\newblock Model-based reinforcement learning via latent-space collocation.
\newblock In \emph{International Conference on Machine Learning}, 2021.

\bibitem[Gelada et~al.(2019)Gelada, Kumar, Buckman, Nachum, and
  Bellemare]{gelada2019deepmdp}
C.~Gelada, S.~Kumar, J.~Buckman, O.~Nachum, and M.~G. Bellemare.
\newblock Deepmdp: Learning continuous latent space models for representation
  learning.
\newblock In \emph{International Conference on Machine Learning}, 2019.

\bibitem[Nguyen et~al.(2021)Nguyen, Shu, Pham, Bui, and
  Ermon]{nguyen2021temporal}
T.~D. Nguyen, R.~Shu, T.~Pham, H.~Bui, and S.~Ermon.
\newblock Temporal predictive coding for model-based planning in latent space.
\newblock In \emph{International Conference on Machine Learning}, 2021.

\bibitem[Okada and Taniguchi(2021)]{okada2021dreaming}
M.~Okada and T.~Taniguchi.
\newblock Dreaming: Model-based reinforcement learning by latent imagination
  without reconstruction.
\newblock In \emph{2021 IEEE International Conference on Robotics and
  Automation (ICRA)}, 2021.

\bibitem[Deng et~al.(2021)Deng, Jang, and Ahn]{deng2021dreamerpro}
F.~Deng, I.~Jang, and S.~Ahn.
\newblock Dreamerpro: Reconstruction-free model-based reinforcement learning
  with prototypical representations.
\newblock \emph{arXiv preprint arXiv:2110.14565}, 2021.

\bibitem[Chen et~al.(2021)Chen, Xie, and He]{chen2021empirical}
X.~Chen, S.~Xie, and K.~He.
\newblock An empirical study of training self-supervised vision transformers.
\newblock In \emph{Proceedings of the IEEE/CVF International Conference on
  Computer Vision}, 2021.

\bibitem[Caron et~al.(2021)Caron, Touvron, Misra, J{\'e}gou, Mairal,
  Bojanowski, and Joulin]{caron2021emerging}
M.~Caron, H.~Touvron, I.~Misra, H.~J{\'e}gou, J.~Mairal, P.~Bojanowski, and
  A.~Joulin.
\newblock Emerging properties in self-supervised vision transformers.
\newblock In \emph{Proceedings of the IEEE/CVF International Conference on
  Computer Vision}, 2021.

\bibitem[Hinton et~al.(2015)Hinton, Vinyals, Dean,
  et~al.]{hinton2015distilling}
G.~Hinton, O.~Vinyals, J.~Dean, et~al.
\newblock Distilling the knowledge in a neural network.
\newblock \emph{arXiv preprint arXiv:1503.02531}, 2015.

\bibitem[Bao et~al.(2021)Bao, Dong, and Wei]{bao2021beit}
H.~Bao, L.~Dong, and F.~Wei.
\newblock Beit: Bert pre-training of image transformers.
\newblock \emph{arXiv preprint arXiv:2106.08254}, 2021.

\bibitem[Li et~al.(2021)Li, Chen, Yang, Li, Zhu, Zhao, Deng, Wu, Zhao, Tang,
  et~al.]{li2021mst}
Z.~Li, Z.~Chen, F.~Yang, W.~Li, Y.~Zhu, C.~Zhao, R.~Deng, L.~Wu, R.~Zhao,
  M.~Tang, et~al.
\newblock Mst: Masked self-supervised transformer for visual representation.
\newblock In \emph{Advances in Neural Information Processing Systems}, 2021.

\bibitem[Feichtenhofer et~al.(2022)Feichtenhofer, Fan, Li, and
  He]{feichtenhofer2022masked}
C.~Feichtenhofer, H.~Fan, Y.~Li, and K.~He.
\newblock Masked autoencoders as spatiotemporal learners.
\newblock \emph{arXiv preprint arXiv:2205.09113}, 2022.

\bibitem[Xie et~al.(2022)Xie, Zhang, Cao, Lin, Bao, Yao, Dai, and
  Hu]{xie2022simmim}
Z.~Xie, Z.~Zhang, Y.~Cao, Y.~Lin, J.~Bao, Z.~Yao, Q.~Dai, and H.~Hu.
\newblock Simmim: A simple framework for masked image modeling.
\newblock In \emph{Proceedings of the IEEE/CVF Conference on Computer Vision
  and Pattern Recognition}, 2022.

\bibitem[Wei et~al.(2022)Wei, Fan, Xie, Wu, Yuille, and
  Feichtenhofer]{wei2022masked}
C.~Wei, H.~Fan, S.~Xie, C.-Y. Wu, A.~Yuille, and C.~Feichtenhofer.
\newblock Masked feature prediction for self-supervised visual pre-training.
\newblock In \emph{Proceedings of the IEEE/CVF Conference on Computer Vision
  and Pattern Recognition}, 2022.

\bibitem[Zhou et~al.(2021)Zhou, Wei, Wang, Shen, Xie, Yuille, and
  Kong]{zhou2021ibot}
J.~Zhou, C.~Wei, H.~Wang, W.~Shen, C.~Xie, A.~Yuille, and T.~Kong.
\newblock ibot: Image bert pre-training with online tokenizer.
\newblock \emph{arXiv preprint arXiv:2111.07832}, 2021.

\bibitem[Xiao et~al.(2021)Xiao, Singh, Mintun, Darrell, Doll{\'a}r, and
  Girshick]{xiao2021early}
T.~Xiao, M.~Singh, E.~Mintun, T.~Darrell, P.~Doll{\'a}r, and R.~Girshick.
\newblock Early convolutions help transformers see better.
\newblock In \emph{Advances in Neural Information Processing Systems}, 2021.

\bibitem[Tassa et~al.(2020)Tassa, Tunyasuvunakool, Muldal, Doron, Liu, Bohez,
  Merel, Erez, Lillicrap, and Heess]{tassa2020dm_control}
Y.~Tassa, S.~Tunyasuvunakool, A.~Muldal, Y.~Doron, S.~Liu, S.~Bohez, J.~Merel,
  T.~Erez, T.~Lillicrap, and N.~Heess.
\newblock dm\_control: Software and tasks for continuous control.
\newblock \emph{arXiv preprint arXiv:2006.12983}, 2020.

\bibitem[Sutton and Barto(2018)]{sutton2018reinforcement}
R.~S. Sutton and A.~G. Barto.
\newblock \emph{Reinforcement learning: An introduction}.
\newblock MIT Press, 2018.

\bibitem[Alemi et~al.(2018)Alemi, Poole, Fischer, Dillon, Saurous, and
  Murphy]{alemi2018fixing}
A.~Alemi, B.~Poole, I.~Fischer, J.~Dillon, R.~A. Saurous, and K.~Murphy.
\newblock Fixing a broken elbo.
\newblock In \emph{International Conference on Machine Learning}, 2018.

\bibitem[Vaswani et~al.(2017)Vaswani, Shazeer, Parmar, Uszkoreit, Jones, Gomez,
  Kaiser, and Polosukhin]{vaswani2017attention}
A.~Vaswani, N.~Shazeer, N.~Parmar, J.~Uszkoreit, L.~Jones, A.~N. Gomez,
  {\L}.~Kaiser, and I.~Polosukhin.
\newblock Attention is all you need.
\newblock In \emph{Advances in Neural Information Processing Systems}, 2017.

\bibitem[Graham et~al.(2021)Graham, El-Nouby, Touvron, Stock, Joulin,
  J{\'e}gou, and Douze]{graham2021levit}
B.~Graham, A.~El-Nouby, H.~Touvron, P.~Stock, A.~Joulin, H.~J{\'e}gou, and
  M.~Douze.
\newblock Levit: a vision transformer in convnet's clothing for faster
  inference.
\newblock In \emph{Proceedings of the IEEE/CVF International Conference on
  Computer Vision}, 2021.

\bibitem[Tassa et~al.(2012)Tassa, Erez, and Todorov]{tassa2012synthesis}
Y.~Tassa, T.~Erez, and E.~Todorov.
\newblock Synthesis and stabilization of complex behaviors through online
  trajectory optimization.
\newblock In \emph{International Conference on Intelligent Robots and Systems},
  2012.

\bibitem[James and Davison(2022)]{james2022q}
S.~James and A.~J. Davison.
\newblock Q-attention: Enabling efficient learning for vision-based robotic
  manipulation.
\newblock \emph{IEEE Robotics and Automation Letters}, 2022.

\bibitem[James et~al.(2022)James, Wada, Laidlow, and Davison]{james2021coarse}
S.~James, K.~Wada, T.~Laidlow, and A.~J. Davison.
\newblock Coarse-to-{F}ine {Q}-attention: Efficient learning for visual robotic
  manipulation via discretisation.
\newblock \emph{Proceedings of the IEEE/CVF Conference on Computer Vision and
  Pattern Recognition}, 2022.

\bibitem[James and Abbeel(2022{\natexlab{a}})]{james2022lpr}
S.~James and P.~Abbeel.
\newblock {C}oarse-to-{F}ine {Q}-attention with {L}earned {P}ath {R}anking.
\newblock \emph{arXiv preprint arXiv:2204.01571}, 2022{\natexlab{a}}.

\bibitem[James and Abbeel(2022{\natexlab{b}})]{james2022tree}
S.~James and P.~Abbeel.
\newblock {C}oarse-to-{F}ine {Q}-attention with {T}ree {E}xpansion.
\newblock \emph{arXiv preprint arXiv:2204.12471}, 2022{\natexlab{b}}.

\bibitem[Tong et~al.(2022)Tong, Song, Wang, and Wang]{tong2022videomae}
Z.~Tong, Y.~Song, J.~Wang, and L.~Wang.
\newblock Videomae: Masked autoencoders are data-efficient learners for
  self-supervised video pre-training.
\newblock In \emph{Advances in Neural Information Processing Systems}, 2022.

\bibitem[Arnab et~al.(2021)Arnab, Dehghani, Heigold, Sun, Lu{\v{c}}i{\'c}, and
  Schmid]{arnab2021vivit}
A.~Arnab, M.~Dehghani, G.~Heigold, C.~Sun, M.~Lu{\v{c}}i{\'c}, and C.~Schmid.
\newblock Vivit: A video vision transformer.
\newblock In \emph{Proceedings of the IEEE/CVF International Conference on
  Computer Vision}, 2021.

\bibitem[Geng et~al.(2022)Geng, Liu, Lee, Schuurams, Levine, and
  Abbeel]{geng2022multimodal}
X.~Geng, H.~Liu, L.~Lee, D.~Schuurams, S.~Levine, and P.~Abbeel.
\newblock Multimodal masked autoencoders learn transferable representations.
\newblock \emph{arXiv preprint arXiv:2205.14204}, 2022.

\bibitem[Bachmann et~al.(2022)Bachmann, Mizrahi, Atanov, and
  Zamir]{bachmann2022multimae}
R.~Bachmann, D.~Mizrahi, A.~Atanov, and A.~Zamir.
\newblock Multimae: Multi-modal multi-task masked autoencoders.
\newblock \emph{arXiv preprint arXiv:2204.01678}, 2022.

\bibitem[Seo et~al.(2022)Seo, Lee, James, and Abbeel]{seo2022reinforcement}
Y.~Seo, K.~Lee, S.~James, and P.~Abbeel.
\newblock Reinforcement learning with action-free pre-training from videos.
\newblock In \emph{International Conference on Machine Learning}, 2022.

\bibitem[Schulman et~al.(2015)Schulman, Moritz, Levine, Jordan, and
  Abbeel]{schulman2015high}
J.~Schulman, P.~Moritz, S.~Levine, M.~Jordan, and P.~Abbeel.
\newblock High-dimensional continuous control using generalized advantage
  estimation.
\newblock \emph{arXiv preprint arXiv:1506.02438}, 2015.

\bibitem[He et~al.(2016)He, Zhang, Ren, and Sun]{he2016deep}
K.~He, X.~Zhang, S.~Ren, and J.~Sun.
\newblock Deep residual learning for image recognition.
\newblock In \emph{Proceedings of the IEEE conference on computer vision and
  pattern recognition}, 2016.

\bibitem[Wu et~al.(2021)Wu, Xiao, Codella, Liu, Dai, Yuan, and
  Zhang]{wu2021cvt}
H.~Wu, B.~Xiao, N.~Codella, M.~Liu, X.~Dai, L.~Yuan, and L.~Zhang.
\newblock Cvt: Introducing convolutions to vision transformers.
\newblock In \emph{Proceedings of the IEEE/CVF International Conference on
  Computer Vision}, 2021.

\bibitem[Dai et~al.(2015)Dai, He, and Sun]{dai2015convolutional}
J.~Dai, K.~He, and J.~Sun.
\newblock Convolutional feature masking for joint object and stuff
  segmentation.
\newblock In \emph{Proceedings of the IEEE conference on computer vision and
  pattern recognition}, pages 3992--4000, 2015.

\bibitem[Tompson et~al.(2015)Tompson, Goroshin, Jain, LeCun, and
  Bregler]{tompson2015efficient}
J.~Tompson, R.~Goroshin, A.~Jain, Y.~LeCun, and C.~Bregler.
\newblock Efficient object localization using convolutional networks.
\newblock In \emph{Proceedings of the IEEE conference on computer vision and
  pattern recognition}, 2015.

\bibitem[Ghiasi et~al.(2018)Ghiasi, Lin, and Le]{ghiasi2018dropblock}
G.~Ghiasi, T.-Y. Lin, and Q.~V. Le.
\newblock Dropblock: A regularization method for convolutional networks.
\newblock \emph{Advances in neural information processing systems}, 31, 2018.

\bibitem[Srivastava et~al.(2014)Srivastava, Hinton, Krizhevsky, Sutskever, and
  Salakhutdinov]{srivastava2014dropout}
N.~Srivastava, G.~Hinton, A.~Krizhevsky, I.~Sutskever, and R.~Salakhutdinov.
\newblock Dropout: a simple way to prevent neural networks from overfitting.
\newblock \emph{The journal of machine learning research}, 2014.

\bibitem[Park et~al.(2021)Park, Seo, Liu, Zhao, Qin, Shin, and
  Liu]{park2021object}
J.~Park, Y.~Seo, C.~Liu, L.~Zhao, T.~Qin, J.~Shin, and T.-Y. Liu.
\newblock Object-aware regularization for addressing causal confusion in
  imitation learning.
\newblock In \emph{Advances in Neural Information Processing Systems}, 2021.

\bibitem[Van Den~Oord et~al.(2017)Van Den~Oord, Vinyals, et~al.]{van2017neural}
A.~Van Den~Oord, O.~Vinyals, et~al.
\newblock Neural discrete representation learning.
\newblock In \emph{Advances in Neural Information Processing Systems}, 2017.

\bibitem[Jaderberg et~al.(2017)Jaderberg, Mnih, Czarnecki, Schaul, Leibo,
  Silver, and Kavukcuoglu]{jaderberg2016reinforcement}
M.~Jaderberg, V.~Mnih, W.~M. Czarnecki, T.~Schaul, J.~Z. Leibo, D.~Silver, and
  K.~Kavukcuoglu.
\newblock Reinforcement learning with unsupervised auxiliary tasks.
\newblock In \emph{International Conference on Learning Representations}, 2017.

\bibitem[Schwarzer et~al.(2021)Schwarzer, Anand, Goel, Hjelm, Courville, and
  Bachman]{schwarzer2020data}
M.~Schwarzer, A.~Anand, R.~Goel, R.~D. Hjelm, A.~Courville, and P.~Bachman.
\newblock Data-efficient reinforcement learning with self-predictive
  representations.
\newblock In \emph{International Conference on Learning Representations}, 2021.

\bibitem[Yu et~al.(2021)Yu, Lan, Zeng, Feng, Zhang, and
  Chen]{yu2021playvirtual}
T.~Yu, C.~Lan, W.~Zeng, M.~Feng, Z.~Zhang, and Z.~Chen.
\newblock Playvirtual: Augmenting cycle-consistent virtual trajectories for
  reinforcement learning.
\newblock In \emph{Advances in Neural Information Processing Systems}, 2021.

\bibitem[Yu et~al.(2022)Yu, Zhang, Lan, Chen, and Lu]{yu2022mask}
T.~Yu, Z.~Zhang, C.~Lan, Z.~Chen, and Y.~Lu.
\newblock Mask-based latent reconstruction for reinforcement learning.
\newblock \emph{arXiv preprint arXiv:2201.12096}, 2022.

\bibitem[Castro(2020)]{castro2020scalable}
P.~S. Castro.
\newblock Scalable methods for computing state similarity in deterministic
  markov decision processes.
\newblock In \emph{Proceedings of the AAAI Conference on Artificial
  Intelligence}, 2020.

\bibitem[Zhang et~al.(2021)Zhang, McAllister, Calandra, Gal, and
  Levine]{zhang2020learning}
A.~Zhang, R.~McAllister, R.~Calandra, Y.~Gal, and S.~Levine.
\newblock Learning invariant representations for reinforcement learning without
  reconstruction.
\newblock In \emph{International Conference on Learning Representations}, 2021.

\bibitem[Oord et~al.(2018)Oord, Li, and Vinyals]{oord2018representation}
A.~v.~d. Oord, Y.~Li, and O.~Vinyals.
\newblock Representation learning with contrastive predictive coding.
\newblock In \emph{Advances in Neural Information Processing Systems}, 2018.

\bibitem[Anand et~al.(2019)Anand, Racah, Ozair, Bengio, C{\^o}t{\'e}, and
  Hjelm]{anand2019unsupervised}
A.~Anand, E.~Racah, S.~Ozair, Y.~Bengio, M.-A. C{\^o}t{\'e}, and R.~D. Hjelm.
\newblock Unsupervised state representation learning in atari.
\newblock In \emph{Advances in Neural Information Processing Systems}, 2019.

\bibitem[Mazoure et~al.(2020)Mazoure, Combes, Doan, Bachman, and
  Hjelm]{mazoure2020deep}
B.~Mazoure, R.~T.~d. Combes, T.~Doan, P.~Bachman, and R.~D. Hjelm.
\newblock Deep reinforcement and infomax learning.
\newblock In \emph{Advances in Neural Information Processing Systems}, 2020.

\bibitem[Srinivas et~al.(2020)Srinivas, Laskin, and Abbeel]{srinivas2020curl}
A.~Srinivas, M.~Laskin, and P.~Abbeel.
\newblock Curl: Contrastive unsupervised representations for reinforcement
  learning.
\newblock In \emph{Internatial Conference on Machine Learning}, 2020.

\bibitem[Liu and Abbeel(2021)]{liu2021behavior}
H.~Liu and P.~Abbeel.
\newblock Behavior from the void: Unsupervised active pre-training.
\newblock \emph{arXiv preprint arXiv:2103.04551}, 2021.

\bibitem[Yarats et~al.(2021)Yarats, Fergus, Lazaric, and
  Pinto]{yarats2021reinforcement}
D.~Yarats, R.~Fergus, A.~Lazaric, and L.~Pinto.
\newblock Reinforcement learning with prototypical representations.
\newblock In \emph{International Conference on Machine Learning}, 2021.

\bibitem[Nair et~al.(2022)Nair, Rajeswaran, Kumar, Finn, and
  Gupta]{nair2022r3m}
S.~Nair, A.~Rajeswaran, V.~Kumar, C.~Finn, and A.~Gupta.
\newblock R3m: A universal visual representation for robot manipulation.
\newblock \emph{arXiv preprint arXiv:2203.12601}, 2022.

\bibitem[Parisi et~al.(2022)Parisi, Rajeswaran, Purushwalkam, and
  Gupta]{parisi2022unsurprising}
S.~Parisi, A.~Rajeswaran, S.~Purushwalkam, and A.~Gupta.
\newblock The unsurprising effectiveness of pre-trained vision models for
  control.
\newblock \emph{arXiv preprint arXiv:2203.03580}, 2022.

\bibitem[Minderer et~al.(2019)Minderer, Sun, Villegas, Cole, Murphy, and
  Lee]{minderer2019unsupervised}
M.~Minderer, C.~Sun, R.~Villegas, F.~Cole, K.~P. Murphy, and H.~Lee.
\newblock Unsupervised learning of object structure and dynamics from videos.
\newblock In \emph{Advances in Neural Information Processing Systems}, 2019.

\bibitem[Manuelli et~al.(2020)Manuelli, Li, Florence, and
  Tedrake]{manuelli2020keypoints}
L.~Manuelli, Y.~Li, P.~Florence, and R.~Tedrake.
\newblock Keypoints into the future: Self-supervised correspondence in
  model-based reinforcement learning.
\newblock \emph{arXiv preprint arXiv:2009.05085}, 2020.

\bibitem[Lambeta et~al.(2020)Lambeta, Chou, Tian, Yang, Maloon, Most, Stroud,
  Santos, Byagowi, Kammerer, et~al.]{lambeta2020digit}
M.~Lambeta, P.-W. Chou, S.~Tian, B.~Yang, B.~Maloon, V.~R. Most, D.~Stroud,
  R.~Santos, A.~Byagowi, G.~Kammerer, et~al.
\newblock Digit: A novel design for a low-cost compact high-resolution tactile
  sensor with application to in-hand manipulation.
\newblock \emph{IEEE Robotics and Automation Letters}, 2020.

\bibitem[Das et~al.(2021)Das, Bechtle, Davchev, Jayaraman, Rai, and
  Meier]{das2021model}
N.~Das, S.~Bechtle, T.~Davchev, D.~Jayaraman, A.~Rai, and F.~Meier.
\newblock Model-based inverse reinforcement learning from visual
  demonstrations.
\newblock In \emph{Conference on Robot Learning}, 2021.

\bibitem[Yarats et~al.(2021)Yarats, Zhang, Kostrikov, Amos, Pineau, and
  Fergus]{yarats2019improving}
D.~Yarats, A.~Zhang, I.~Kostrikov, B.~Amos, J.~Pineau, and R.~Fergus.
\newblock Improving sample efficiency in model-free reinforcement learning from
  images.
\newblock In \emph{Proceedings of the AAAI Conference on Artificial
  Intelligence}, 2021.

\bibitem[Shelhamer et~al.(2016)Shelhamer, Mahmoudieh, Argus, and
  Darrell]{shelhamer2016loss}
E.~Shelhamer, P.~Mahmoudieh, M.~Argus, and T.~Darrell.
\newblock Loss is its own reward: Self-supervision for reinforcement learning.
\newblock \emph{arXiv preprint arXiv:1612.07307}, 2016.

\bibitem[Yarats et~al.(2021)Yarats, Fergus, Lazaric, and
  Pinto]{yarats2021mastering}
D.~Yarats, R.~Fergus, A.~Lazaric, and L.~Pinto.
\newblock Mastering visual continuous control: Improved data-augmented
  reinforcement learning.
\newblock \emph{arXiv preprint arXiv:2107.09645}, 2021.

\bibitem[Laskin et~al.(2020)Laskin, Lee, Stooke, Pinto, Abbeel, and
  Srinivas]{laskin2020reinforcement}
M.~Laskin, K.~Lee, A.~Stooke, L.~Pinto, P.~Abbeel, and A.~Srinivas.
\newblock Reinforcement learning with augmented data.
\newblock In \emph{Advances in Neural Information Processing Systems}, 2020.

\bibitem[Kostrikov et~al.(2021)Kostrikov, Yarats, and
  Fergus]{kostrikov2020image}
I.~Kostrikov, D.~Yarats, and R.~Fergus.
\newblock Image augmentation is all you need: Regularizing deep reinforcement
  learning from pixels.
\newblock In \emph{International Conference on Learning Representations}, 2021.

\bibitem[Xiao et~al.(2022)Xiao, Radosavovic, Darrell, and
  Malik]{xiao2022masked}
T.~Xiao, I.~Radosavovic, T.~Darrell, and J.~Malik.
\newblock Masked visual pre-training for motor control.
\newblock \emph{arXiv preprint arXiv:2203.06173}, 2022.

\end{thebibliography}

\newpage
\appendix
\onecolumn

\begin{center}{\bf {\LARGE Appendix}}
\end{center}

\section{Behavior Learning}
\label{appendix:behavior_learning}
We utilize the actor-critic learning scheme of DreamerV2~\citep{hafner2020mastering}. Specifically, we introduce a stochastic actor and a deterministic critic as below:
\begin{gather}
\begin{aligned}
&\text{Actor:} &&\hat{a}_{t} \sim p_{\psi}(\hat{a}_{t}\,|\,\hat{s}_{t}) &&\text{Critic:} &&v_{\xi}(\hat{s}_{t}) \approx \mathbb{E}_{p_{\theta}}\left[\textstyle\sum_{i \leq t}\gamma^{i - t} \hat{r}_{i}\right],
\label{eq:actor_critic}
\end{aligned}
\end{gather}
where \{$\hat{s}_{t}$, $\hat{a}_{t}$, $\hat{r}_{t}$\} is imagined future states, actions, and rewards which are recursively obtained by conditioning on a initial state $\hat{s}_{0}$ and utilizing the transition model and the reward model in~\cref{eq:dreamer_world_model}, and the actor in~\cref{eq:actor_critic}.
Note that the initial state $\hat{s}_{0}$ is the model state obtained from the representation model in~\cref{eq:dreamer_world_model} using the samples from the replay buffer.
Then the critic is trained to regress the $\lambda$-target~\citep{sutton2018reinforcement,schulman2015high} as follows:
\begin{align}
    &\mathcal{L}^{\texttt{critic}}(\xi)\doteq\mathbb{E}_{p_{\theta}}\left[\sum^{H-1}_{t=1} \frac{1}{2} \left(v_{\xi}(\hat{s}_{t}) - \text{sg}(V_{t}^{\lambda})\right)^{2}\right],
    \label{eq:critic_loss}\\
    &V_{t}^{\lambda}\doteq \hat{r}_{t} + \gamma
    \begin{cases}
      (1 - \lambda)v_{\xi}(\hat{s}_{t+1})+\lambda V_{t+1}^{\lambda} & \text{if}\ t<H \\
      v_{\xi}(\hat{s}_{H}) & \text{if}\ t=H,
    \end{cases}
    \label{eq:lambda_return}
\end{align}
where $\text{sg}$ is a stop gradient function.
Then we train the actor that maximizes the imagined return by back propagating the gradients through the learned world models as follows:
\begin{align}
    \mathcal{L}^{\texttt{actor}}(\psi)\doteq \mathbb{E}_{p_{\theta}} \left[-V_{t}^{\lambda} - \eta\,\text{H}\left[a_{t}|\hat{s}_{t}\right] \right],
    \label{eq:actor_loss}
\end{align}
where the entropy of actor $\,\text{H}\left[a_{t}|\hat{s}_{t}\right]$ is maximized to encourage exploration, and $\eta$ is a hyperparameter that adjusts the strength of entropy regularization.
We refer to \citet{hafner2020mastering} for more details.

\section{Extended Qualitative Analysis}
\label{appendix:qualitative_analysis}

We provide our qualitative analysis in videos on our project website:
\begin{align*}
    \text{\url{https://sites.google.com/view/mwm-rl}}
\end{align*}
which contains videos for (i) reconstructions from masked autoencoders (MAE)~\citep{he2021masked} and (ii) predictions from latent dynamics models.
To be self-contained, we also provide reconstructions from masked autoencoders with images in~\cref{appendix:qualitative_analysis_mae}.

\subsection{Reconstructions from Masked Autoencoders}
\label{appendix:qualitative_analysis_mae}

\begin{figure*} [ht] \centering
    \includegraphics[width=0.99\textwidth]{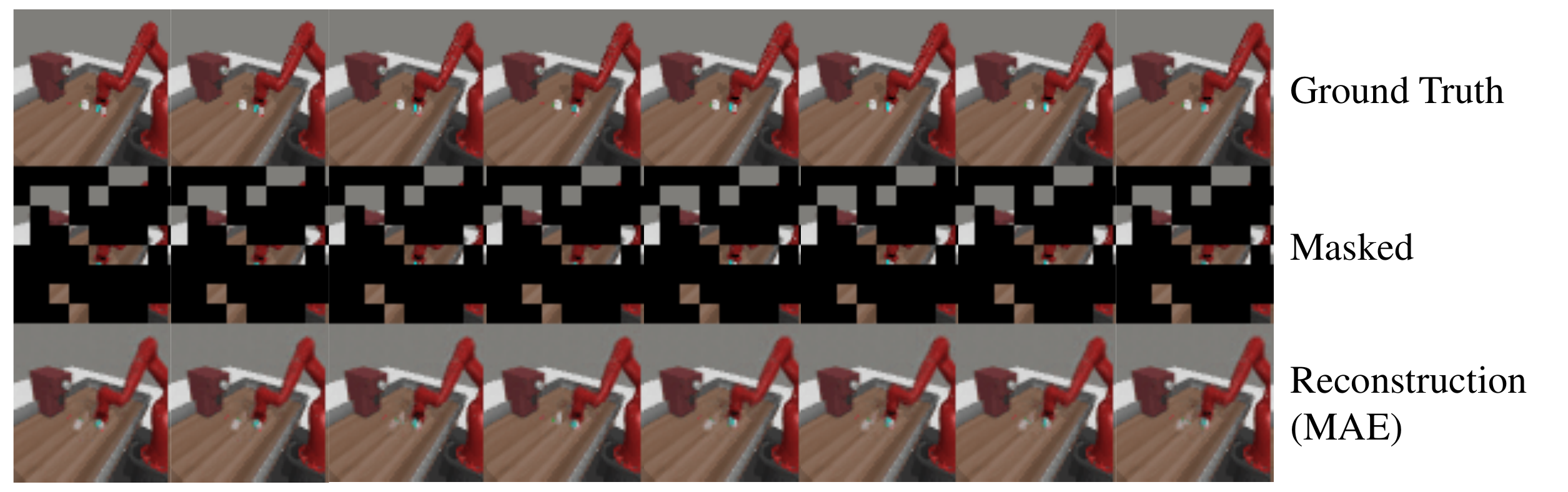}
    \includegraphics[width=0.99\textwidth]{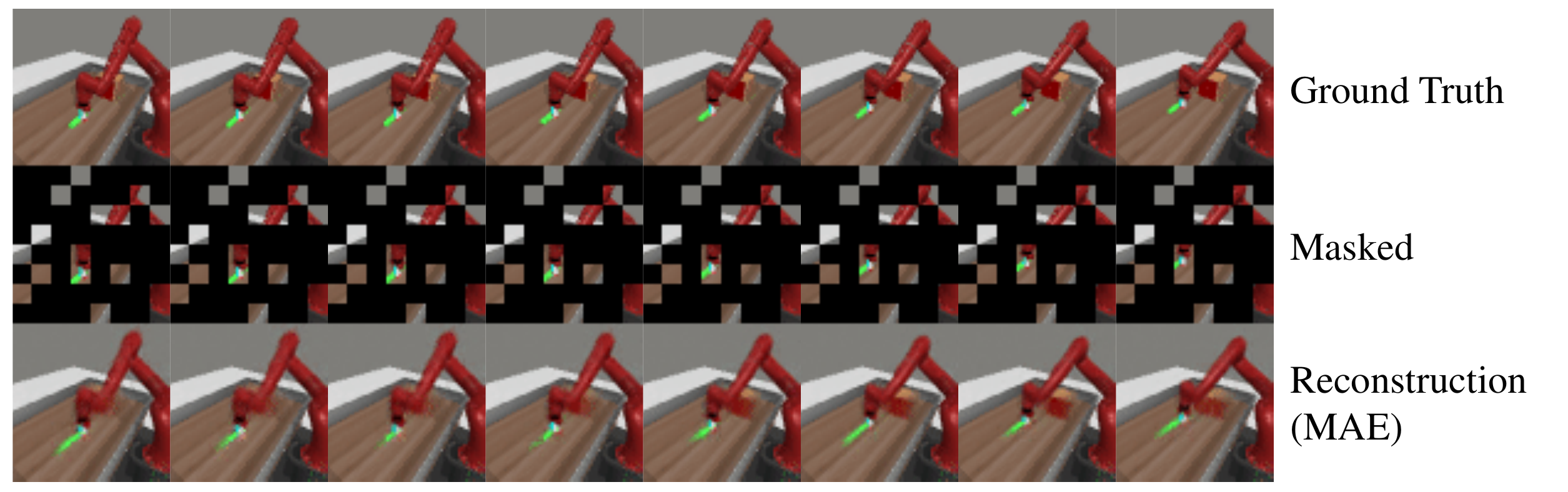}
    \caption{
    Frames reconstructed with the masked autoencoders (MAE)~\citep{he2021masked} trained on Meta-world (Top) Coffee Pull and (Bottom) Peg Insert Side. We find that reconstructions are not capturing the detailed object positions within patches. Best viewed as video provided in our website.
    }
    \label{fig:qualitative_example_gif_reconstruction}
\end{figure*}

In this section, we provide motivating examples for introducing convolutional feature masking. Specifically, we provide reconstructions from MAE~\citep{he2021masked} trained on Coffee-Pull and Peg-Insert-Side tasks from Meta-world~\citep{yu2020meta} in~\cref{fig:qualitative_example_gif_reconstruction}.
We find that reconstruction with pixel patch masking can be an extremely difficult objective, which makes it difficult for the model to learn the fine-grained details such as object positions.
For instance, in~\cref{fig:qualitative_example_gif_reconstruction}, MAE struggles to predict the position of objects (\textit{e.g.,} a cup or a block) within masked patches, making it difficult to learn such details.

\section{Extended Related Work}
\label{appendix:extended_related_work}
\paragraph{Vision transformers with early convolution}
Introducing convolutional layers into a ViT architecture is not new. \citet{dosovitskiy2020image} investigated a hybrid ViT architecture that utilizes a modified version of ResNet~\citep{he2016deep} to obtain a convolutional feature map.
This has been followed by a series of works that investigate the architecture design to introduce convolutions for improved performance~\citep{graham2021levit,wu2021cvt}.
While these works mostly consider deep convolutional networks to maximize the performance on downstream tasks, we introduce a lightweight convolution stem consisting of a few convolution layers, following the design of \citet{xiao2021early}.
This is because our motivation for introducing the convolution stem is to avoid the pitfall of reconstruction objective with masked pixel patches, but not to investigate the optimal hybrid ViT architecture that maximizes the performance.

\paragraph{Convolutional feature masking}
In the context of semantic segmentation, \citet{dai2015convolutional} proposed to utilize convolutional feature masking instead of pixel masking. But this differs in that their goal is to utilize the masks as inputs to classifiers, unlike our approach that drops convolutional features with masks.
More related to our work is the approaches that mask out convolutional features as a regularization technique~\citep{tompson2015efficient,ghiasi2018dropblock}.
For instance, \citet{tompson2015efficient} demonstrated that masking out entire channels for a specific feature from a convolutional feature map can be more effective than Dropout~\citep{srivastava2014dropout} that masks randomly sampled channels.
\citet{ghiasi2018dropblock} further developed this idea by proposing DropBlock that masks contiguous region of a feature map.
In the context of visual control, \citet{park2021object} trained a VQ-VAE~\citep{van2017neural} and proposed to drop convolutional features corresponding to randomly sampled discrete latent codes.
Our work extends the idea of masking out convolutional features to self-supervised learning with a ViT and demonstrates its effectiveness for representation learning in visual model-based RL.

\paragraph{Unsupervised representation learning for visual control}
Following the work of \citet{jaderberg2016reinforcement} that demonstrated the effectiveness of auxiliary unsupervised objectives for RL, a variety of unsupervised learning objectives have been studied, including future latent reconstruction~\citep{gelada2019deepmdp,schwarzer2020data,yu2021playvirtual,yu2022mask}, bisimulation~\citep{castro2020scalable,zhang2020learning}, contrastive learning~\citep{oord2018representation,anand2019unsupervised,mazoure2020deep,srinivas2020curl,liu2021behavior,deng2021dreamerpro,yarats2021reinforcement,okada2021dreaming,nair2022r3m,parisi2022unsurprising}, keypoint extraction~\citep{minderer2019unsupervised,manuelli2020keypoints,lambeta2020digit,das2021model}, world model learning~\citep{watter2015embed,hafner2019learning,zhang2019solar,seo2022reinforcement} and reconstruction~\citep{yarats2019improving,shelhamer2016loss}.
Recent approaches have also demonstrated that simple data augmentations can sometimes be effective even without such representation learning objectives~\citep{yarats2021mastering,laskin2020reinforcement,kostrikov2020image}.
The work closest to ours is \citet{xiao2022masked}, which demonstrated that frozen representations from MAE pre-trained on real-world videos can be used for training RL agents on visual manipulation tasks.
Our work differs in that we demonstrate that training a self-supervised ViT with reconstruction and convolutional feature masking can be more effective for visual control tasks when compared to MAE that masks pixel patches.
We also note that our work is orthogonal to \citet{xiao2022masked} in that our framework can also initialize the autoencoder with parameters pre-trained using real-world videos.

\clearpage

\section{Pseudocode}
\label{appendix:pseudocode}
For clarity, we define the optimization objectives for autoencoder and latent dynamics model, and describe the pseudocode for our method. Specifically, given a random batch $\{(o_{j}, r_{j}, a_{j})\}_{j=1}^{B}$, visual representation learning and dynamics learning objectives are defined as:
\begin{align}
    &\mathcal{L}^{\texttt{vis}}(\phi) = \frac{1}{B}\sum_{j=1}^{B}\Big(-\ln p_{\phi}(o_{j}|z^{c,m}_{j}) -\ln p_{\phi}(r_{j}|z^{c,m}_{j})\Big)\label{eq:ours_vis_objective}\\
    &\mathcal{L}^{\texttt{dyn}}(\theta) = \frac{1}{B}\sum_{j=1}^{B}\Big(-\ln p_{\theta}(z^{c,0}_{j}|s_{j})-\ln p_{\theta}(r_{j}|s_{j})\label{eq:ours_dyn_objective}\\ &\quad\quad\quad\quad\quad\quad\quad\quad+ \beta\,\text{KL}\left[ q_{\theta}(s_{j}|s_{j-1},a_{j-1},z^{c,0}_{j}) \,\Vert\,  p_{\theta}(\hat{s}_{j}|s_{j-1},a_{j-1}) \right]\Big)\nonumber
\end{align}

\begin{algorithm}[h]
\caption{Masked World Models} \label{alg:training}
\begin{algorithmic}[1]
\STATE Initialize parameters of autoencoder $\phi$, latent dynamics model $\theta$, actor $\psi$, and critic $\xi$
\STATE Initialize replay buffer $\mathcal{B} \leftarrow \emptyset$
\FOR{each timestep $t$}
\STATE {{\textsc{// Collect transitions}}}
\STATE Get autoencoder representation $z_{t}^{c, 0}$
\STATE Update model state $s_{t} \sim q_{\theta}(s_{t}|s_{t-1},a_{t-1},z_{t}^{c, 0})$
\STATE Sample action $a_{t} \sim p_{\psi}(a_{t}|s_{t})$
\STATE Add transition to replay buffer $\mathcal{B} \leftarrow \mathcal{B} \cup \{(o_{t}, a_{t}, r_{t})\}$\vspace{0.05in}
\STATE {{\textsc{// Visual representation learning with reward prediction}}}
\STATE Sample random minibatch $\{(o,r)\} \sim \mathcal{B}$
\STATE Update autoencoder by minimizing $\mathcal{L}^{\texttt{vis}}(\phi)$ \vspace{0.05in}
\STATE {{\textsc{// Dynamics learning}}}
\STATE Sample random minibatch $\{(o,r,a)\} \sim \mathcal{B}$
\STATE Update latent dynamics model by minimizing $\mathcal{L}^{\texttt{dyn}}(\theta)$ and obtain states $\{s\}$ \vspace{0.05in}
\STATE {{\textsc{// Actor Critic learning}}}
\STATE Imagine future rollouts $\{\hat{s}, \hat{a}, \hat{r}\}$ from $\{s\}$ using latent dynamics model and actor
\STATE Update actor by minimizing $\mathcal{L}^{\texttt{actor}}(\psi)$
\STATE Update critic by minimizing $\mathcal{L}^{\texttt{critic}}(\xi)$
\ENDFOR
\end{algorithmic}
\end{algorithm}

\section{Architecture Details}
\label{appendix:architecture_details}
\subsection{Autoencoder}
\paragraph{Convolution stem and masking}
We use visual observations of $64\times64\times3$.
For the convolution stem,
similar to the design of \citet{xiao2021early},
we stack 3 convolution layers with the kernel size of $4 \times 4$ and stride 2, followed by a convolution layer with the kernel size of $1 \times 1$.
This convolution stem processes $o_{t}$ into $8 \times 8 \times 256$, which has the same spatial shape of $8 \times 8$ when we use the patchify stem with patch size of $8 \times 8$.
Then a masking is applied with a masking ratio of $m=0.75$.

\paragraph{ViT encoder and decoder}
We use a 4-layer ViT encoder and a 3-layer ViT decoder, which are implemented using tfimm\footnote{\url{https://github.com/martinsbruveris/tensorflow-image-models}} library.
The ViT encoder concatenates class token with un-masked convolutional features, embeds inputs into 256-dimensional vectors, and processes them through Transformer layers.
Then the ViT decoder takes outputs from the encoder and concatenate learnable mask tokens into them.
Here, we use the same learnable mask token for reward prediction, which can be discriminated from other mask tokens because it gets different positional encoding.
Finally, two linear output heads for predicting pixels and rewards are used to generate predictions.
Unlike MAE, we compute the loss on entire pixels because we do not apply masking to pixels.

\paragraph{Initialization with warm-up schedule} We initialize parameters of the autoencoder using $5000$ gradients steps with a linear warm-up schedule over initial $2500$ steps using the samples collected from initial random exploration.
We find this improves sample-efficiency on relatively easy tasks, but does not make significant difference on complex tasks.
This is because better visual representations are used for learning latent dynamics models from the beginning.
However, we also observe that this initialization is not required when we update the parameters in a more short interval (\textit{e.g.,} every 2 timesteps instead of 5 timesteps), because the autoencoder can be trained quickly without introducing such an initialization period.
In our experiments, we use the initialization scheme and update the parameters every 5 timesteps for faster experimentation.

\subsection{Latent Dynamics Model}
\label{appendix:architecture_details_latent}

\paragraph{Architecture} Our model is built upon the discrete latent dynamics model introduced in DreamerV2.
Inputs to our model are representations $z_{t}^{c, 0}$ from the autoencoder, which are of shape $8 \times 8 \times 256$ obtained by processing the visual observations through the convolution stem and ViT encoder of our autoencoder without masking (\textit{i.e.,} $m = 0$).
Because our dynamics model does not take visual observations as inputs, we do not utilize CNN encoder and decoder as in the original architecture.
Instead, we introduce a shallow 2-layer ViT encoder and decoder with the embedding size of 128, which takes $z_{t}^{c, 0}$ as inputs.
Following \citet{seo2022reinforcement}, we increase the hidden size of dense
layers and the model state dimension from 200 to 1024.

\paragraph{Prediction visualization}
While the latent dynamics model is not trained directly to reconstruct raw pixels, its predictions can still be used for visualizing the open-loop predictions.
This is because it is trained to reconstruct $z_{t}^{c, 0}$, which can be processed through the ViT decoder of the autoencoder.
We use this scheme for visualizing the predictions from the model in~\cref{fig:qualitative_analysis_prediction}.

\section{Experiments Details}
\label{appendix:experiments_details}

\paragraph{Meta-world experiments}
In order to use a single camera viewpoint consistently over all 50 tasks, we use the modified $\texttt{corner2}$ camera viewpoint for all tasks.
Specifically, we adjusted the camera position with \texttt{env.model.cam\_pos[2][:]=[0.75, 0.075, 0.7]}, rendering visual observations as in~\cref{fig:example} which enables us to solve non-zero success rate on all tasks. Maximum episode length for Meta-world tasks is 500. We use the action repeat of 2, which we find it easy to solve tasks compared to the action repeat of 1 used in \citet{seo2022reinforcement}.

In our experiments, we classify 50 tasks into $\texttt{easy}$, $\texttt{medium}$, $\texttt{hard}$, and $\texttt{very hard}$ tasks where experiments are run over 500K, 1M, 2M, 3M environments steps with action repeat of 2, respectively.

\begin{table}[h]
\centering
\small
\begin{tabular}{|c|l|}
\hline
Difficulty & \multicolumn{1}{c|}{Tasks}\\ \hline
\texttt{easy}       & \begin{tabular}[c]{@{}l@{}}
Button Press, Button Press Topdown, Button Press Topdown Wall, Button Press Wall,\\
Coffee Button, Dial Turn, Door Close, Door Lock,\\
Door Open, Door Unlock, Drawer Close, Drawer Open,\\
Faucet Close, Faucet Open, Handle Press, Handle Press Side,\\
Handle Pull, Handle Pull Side, Lever Pull, Plate Slide,\\
Plate Slide Back, Plate Slide Back Side, Plate Slide Side, Reach, Reach Wall,\\
Window Close, Window Open, Peg Unplug Side\end{tabular} \\ \hline
\texttt{medium}     & \begin{tabular}[c]{@{}l@{}}
Basketball, Bin Picking, Box Close, Coffee Pull,\\
Coffee Push, Hammer, Peg Insert Side, Push Wall,\\
Soccer, Sweep, Sweep Into\end{tabular}\\ \hline
\texttt{hard}       & Assembly, Hand Insert, Pick Out of Hole, Pick Place, Push, Push Back\\ \hline
\texttt{very hard}  & Shelf Place, Disassemble, Stick Pull, Stick Push, Pick Place Wall\\ \hline
\end{tabular}
\end{table}

\paragraph{RLBench experiments}
We consider two relatively easy tasks (\textit{i.e.,} Reach Target and Push Button) with dense rewards, and utilize an action mode that specifies the delta of joint positions.
Because original RLBench repository does not support shaped rewards for Push Button task, we design a shaped rewards for Push Button following the design of rewards in Reach Target. Specifically, the reward is defined as the sum of (i) the L2 distance of gripper to a button and (ii) the magnitude of the button being pushed.
We set the maximum episode length to 200, and use the action repeat of 2. Because RLBench is designed to be episodic unlike Meta-world, we use the discount prediction scheme in DreamerV2 that introduces a linear head that predicts the termination of each rollout. For visual observations, we use the front RGB observation (see~\cref{fig:example} for an example).

\paragraph{DeepMind Control Suite experiments}
We follow the setup of \citet{hafner2019dream} where the action repeat of 2 is used. We use default camera configurations without modification. We note that direct comparison with the results from \citet{hafner2019dream} is not possible because our experiments are based on DreamerV2 with larger networks (see~\cref{appendix:architecture_details_latent} for architecture details).

\paragraph{Computation} In terms of parameter counts, MWM consists of 25.9M parameters while DreamerV2 consists of 33.2M parameters. However, in terms of training time, MWM takes 5.5 hours for training over 500K environment steps, which is 1.57 times slower than DreamerV2 that takes 3.5 hours, because MWM processes visual observations through low-throughput ViT twice with and without masking.
Given the improvement in final performances and sample-efficiency on complex tasks as demonstrated in our experiments, we note that it is worth spending additional computational costs.

\paragraph{Hyperparameters} We report the hyperparameters used in our experiments in~\cref{tbl:hyperparameters}.

\begin{table}[h]
\caption{Hyperparameters used in our experiments. Unless otherwise specified, we use the same hyperparameters used in DreamerV2~\citep{hafner2020mastering}. DMC is an abbreviation of DeepMind Control Suite.}
\vskip 0.15in
\begin{center}
\begin{tabular}{ll}
\toprule
\textbf{Hyperparameter} & \textbf{Value}  \\
\midrule
Image observation  & $64 \times 64 \times 3$ \\
Image normalization  & Mean: $(0.485, 0.456, 0.406)$, Std: $(0.229, 0.224, 0.225)$ \\
Action repeat    & $2$ \\
Max episode length    & $500$ (Meta-world), $200$ (RLBench), $1000$ (DMC) \\
Early episode termination   & True (RLBench), False otherwise \\
Random exploration   & $5000$ environment steps \\
Reward normalization   & False (DMC), True otherwise \\
World model batch size   & $16$ (DMC), $50$ otherwise \\
World model sequence length  & $50$ \\
World model tradeoff ($\beta$)  & $0.1$ (RLBench), $1.0$ otherwise \\
World model tradeoff free-bits  & $0.1$ (RLBench), $0.01$ otherwise \\
World model ViT encoder size  & $2$ layers, $4$ heads, 128 units \\
World model ViT decoder size  & $2$ layers, $4$ heads, 128 units \\
\midrule
Autoencoder batch size    & $1024$  \\ 
Autoencoder initialization steps    & $5000$  \\
Autoencoder warm-up steps    & $2500$  \\ 
Autoencoder learning rate    & $3\cdot10^{-4}$  \\ 
Autoencoder masking ratio    & $0.75$  \\ 
Autoencoder ViT encoder size    & $4$ layers, 4 heads, 256 units  \\
Autoencoder ViT decoder size    & $3$ layers, 4 heads, 256 units  \\ 
\bottomrule
\end{tabular}
\label{tbl:hyperparameters}
\end{center}
\vskip -0.1in
\end{table} 

\clearpage

\section{Full Meta-world Experiments}
\label{appendix:full_meta_world}
\vspace{-0.1in}
\begin{figure*} [h] \centering
    \includegraphics[width=0.975\textwidth]{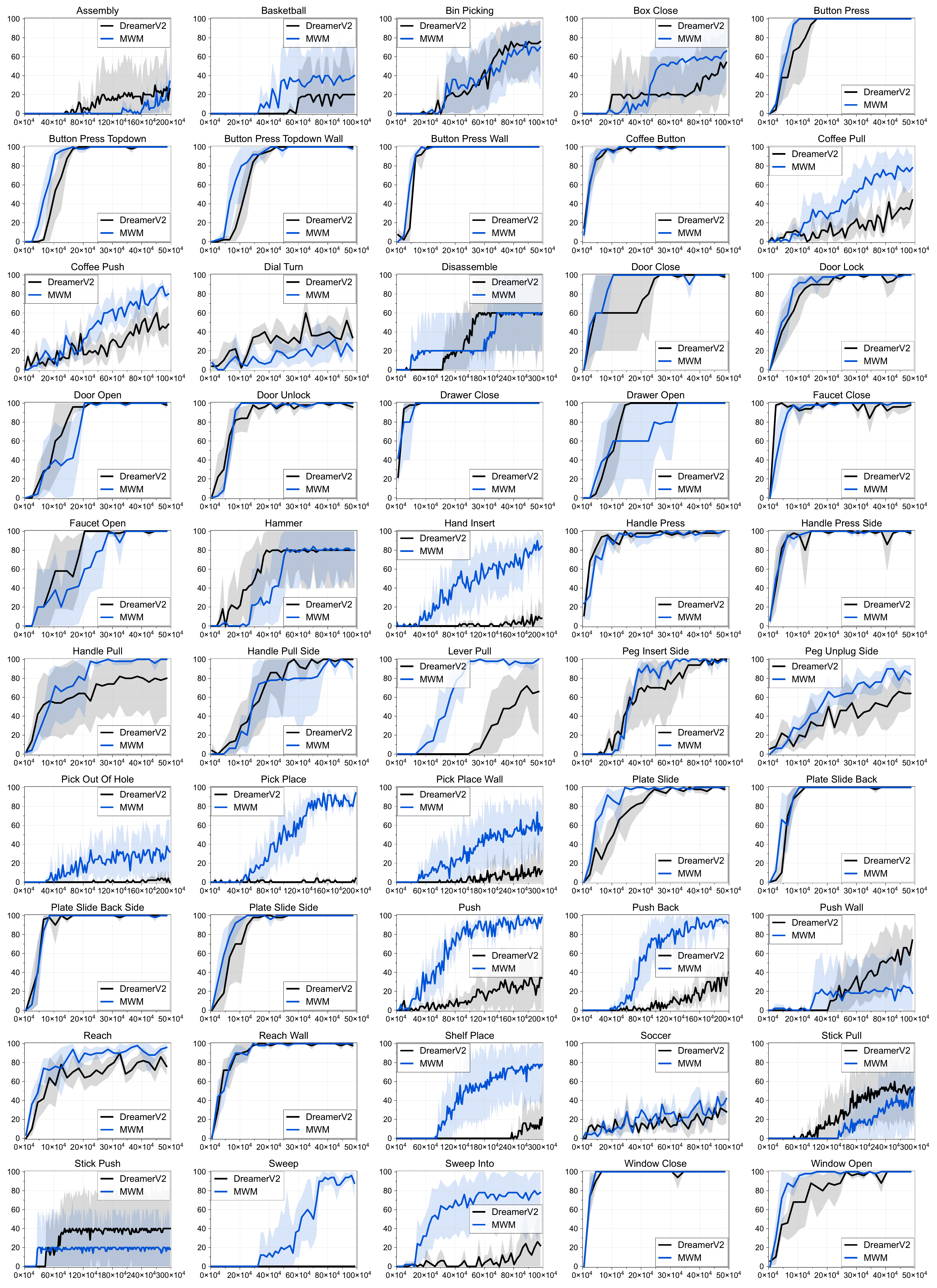}
    \caption{
    Learning curves on 50 visual robotic manipulation tasks from Meta-world as measured on the success rate.
    The solid line and shaded regions represent the mean and bootstrap confidence intervals, respectively, across five runs.}
    \label{fig:metaworld_full}
\end{figure*}
\clearpage

\section{Additional DeepMind Control Suite Experiments}
\label{appendix:additional_dmc}
\vspace{-0.1in}
\begin{figure*} [h!] \centering
    \includegraphics[width=0.99\textwidth]{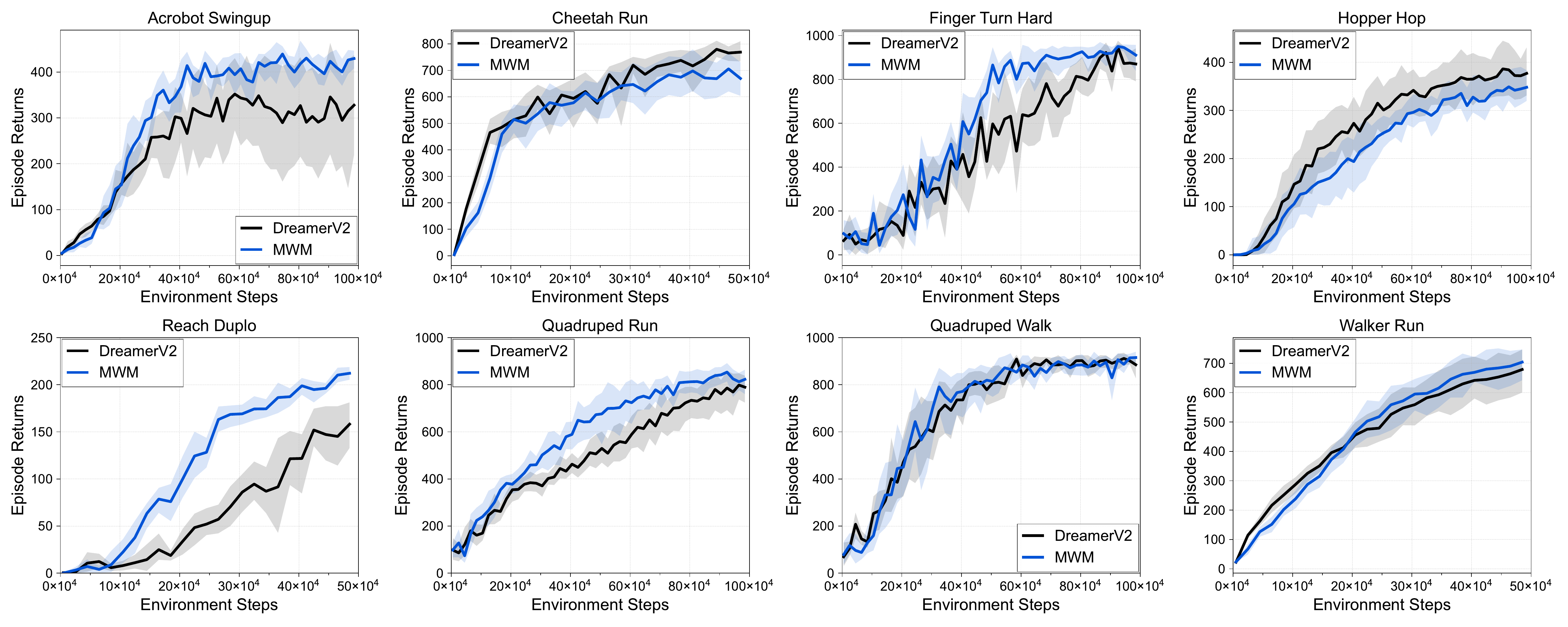}
    \vspace{-0.1in}
    \caption{
    Learning curves on eight visual robot tasks from DeepMind Control Suite as measured on the episode return.
    The solid line and shaded regions represent the mean and bootstrap confidence intervals, respectively, across eight runs.}
    \label{fig:dmc_full}
\end{figure*}

\section{Extended Ablation Study and Analysis}
\label{appendix:full_ablation_analysis}
\begin{figure*} [h] \centering
    \vspace{-0.1in}
    \subfigure[Inputs to world models]
    {
    \includegraphics[width=0.315\textwidth]{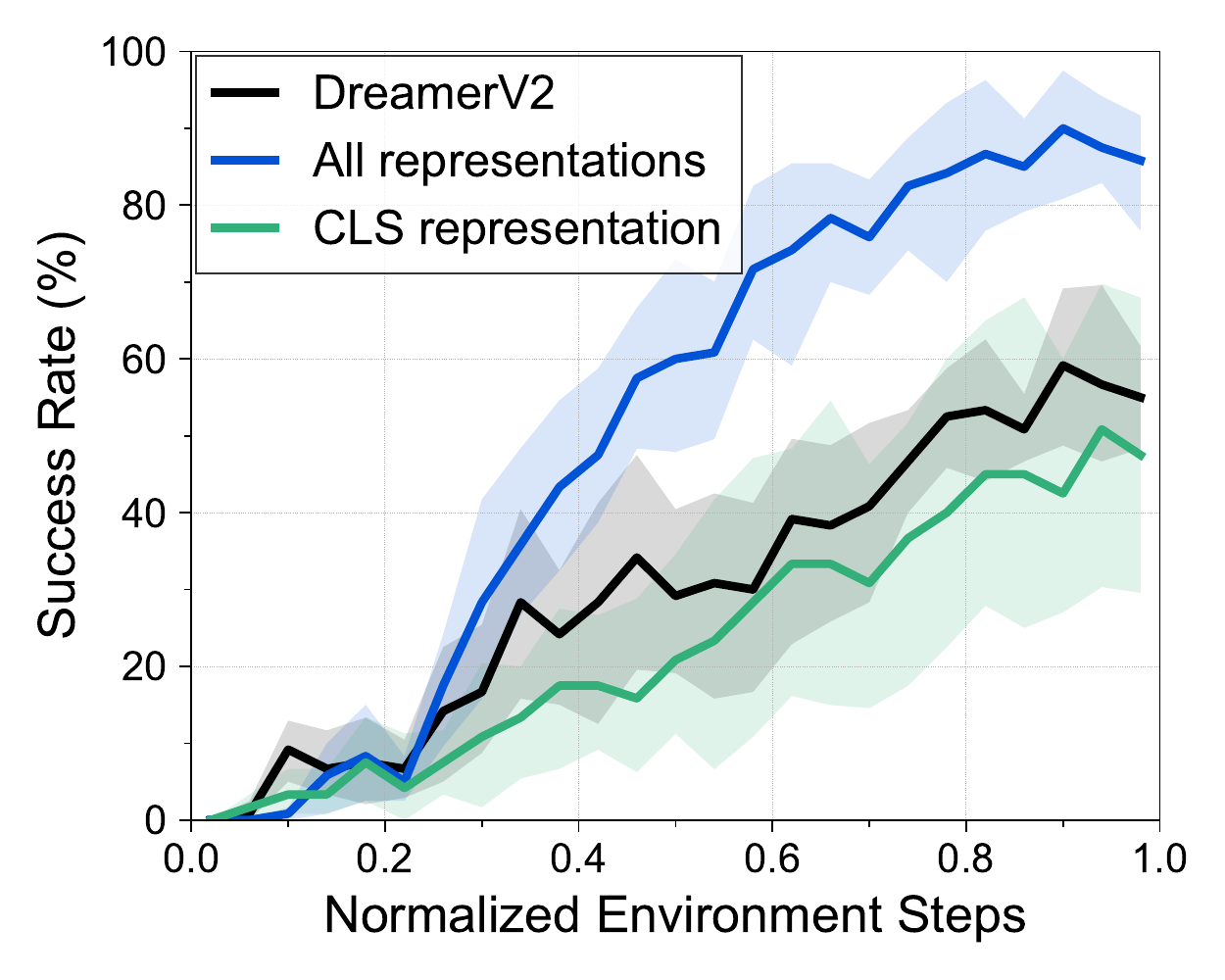}
    \label{fig:metaworld_ablation_class_token}}
    \subfigure[Model size]
    {
    \includegraphics[width=0.315\textwidth]{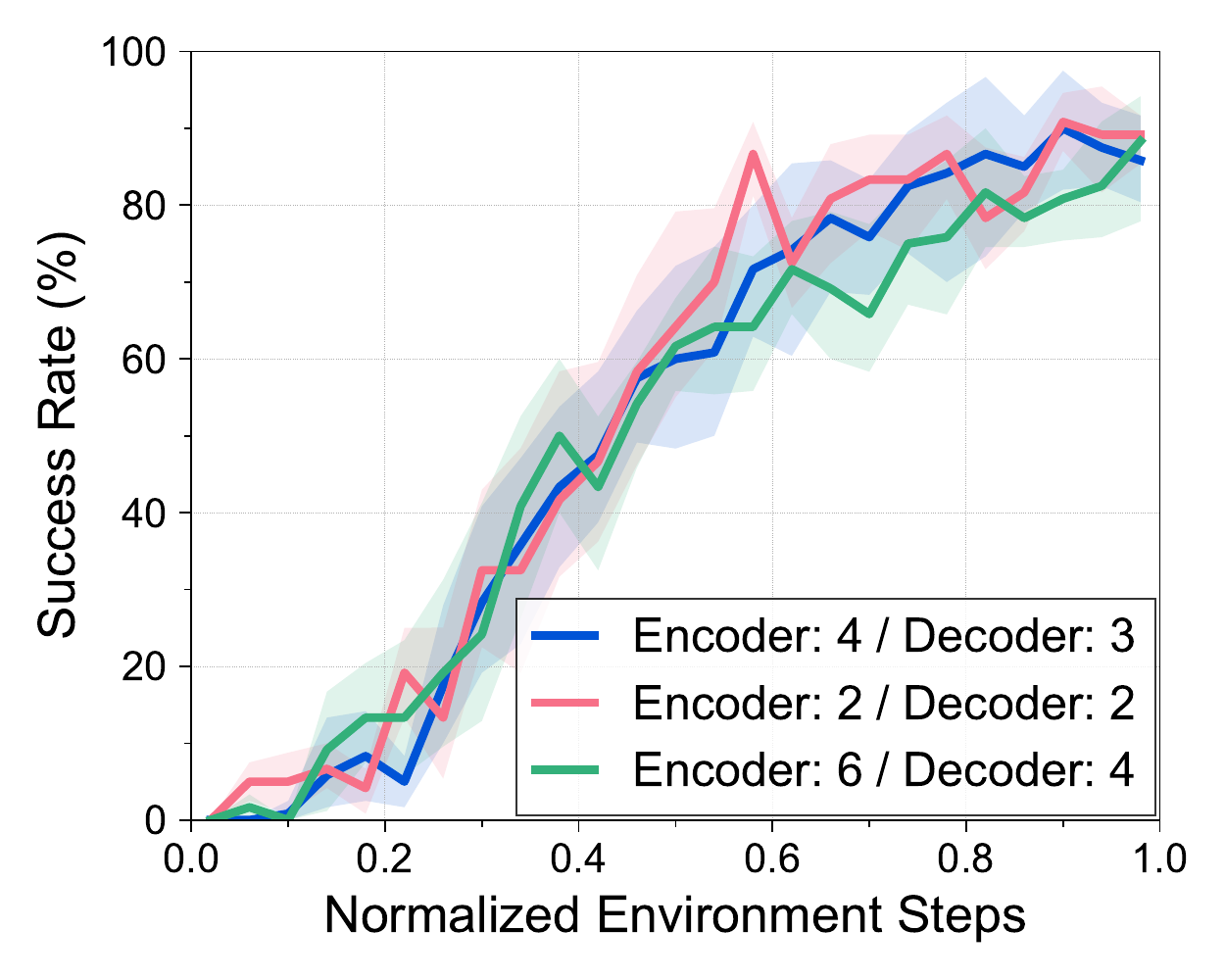}
    \label{fig:metaworld_ablation_model_size}}
    \subfigure[DreamerV2 with ViT]
    {
    \includegraphics[width=0.315\textwidth]{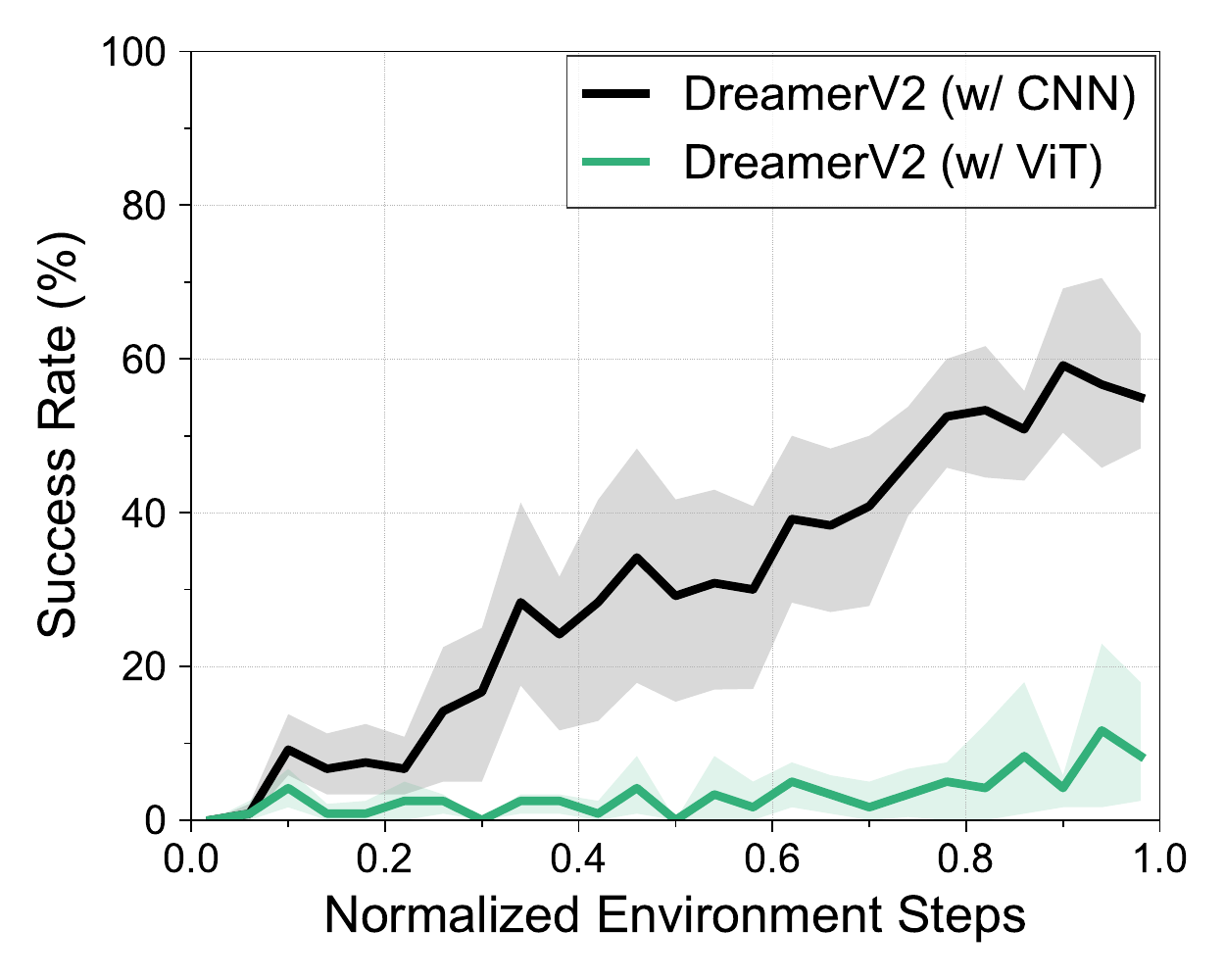}
    \label{fig:metaworld_ablation_vit}}
    \vspace{-0.1in}
    \caption{
    Learning curves on three manipulation tasks from Meta-world that investigate the effect of (a) inputs to world models and (b) autoencoder model sizes. (c) We also report the performance of DreamerV2 with CNNs and ViT.
    The solid line and shaded regions represent the mean and stratified bootstrap confidence interval across 12 runs.
    }
    \label{fig:metaworld_additional_ablation}
\end{figure*}

\begin{figure*} [ht] \centering
    \includegraphics[width=0.96\textwidth]{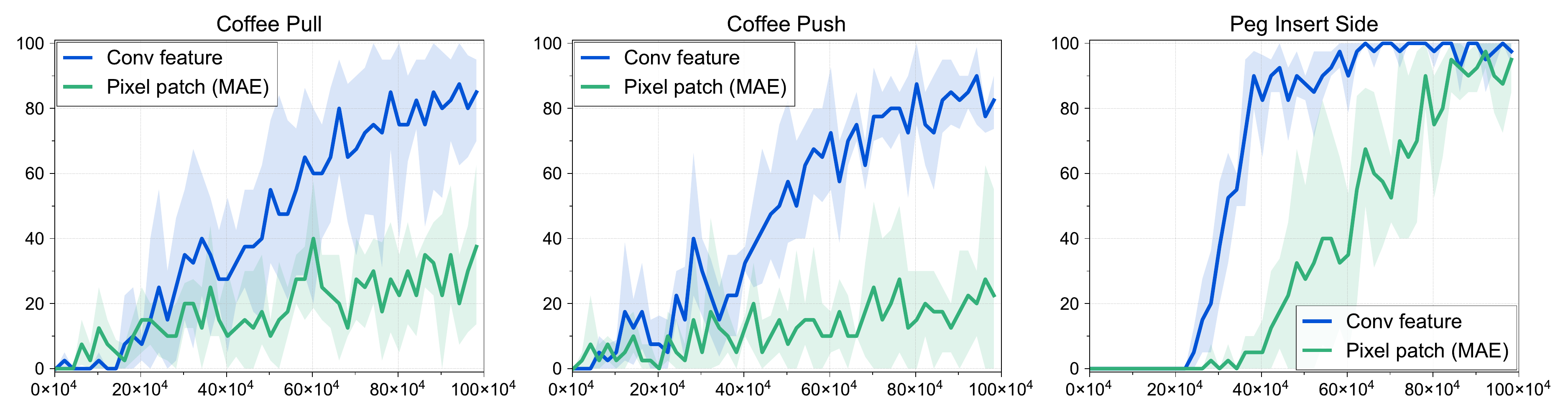}
    \includegraphics[width=0.96\textwidth]{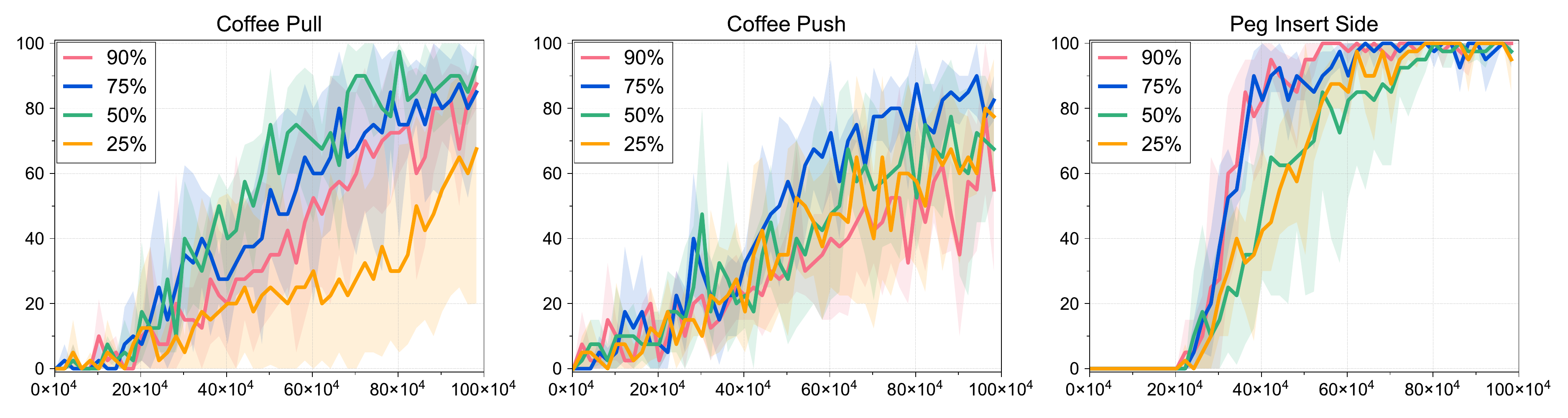}
    \includegraphics[width=0.96\textwidth]{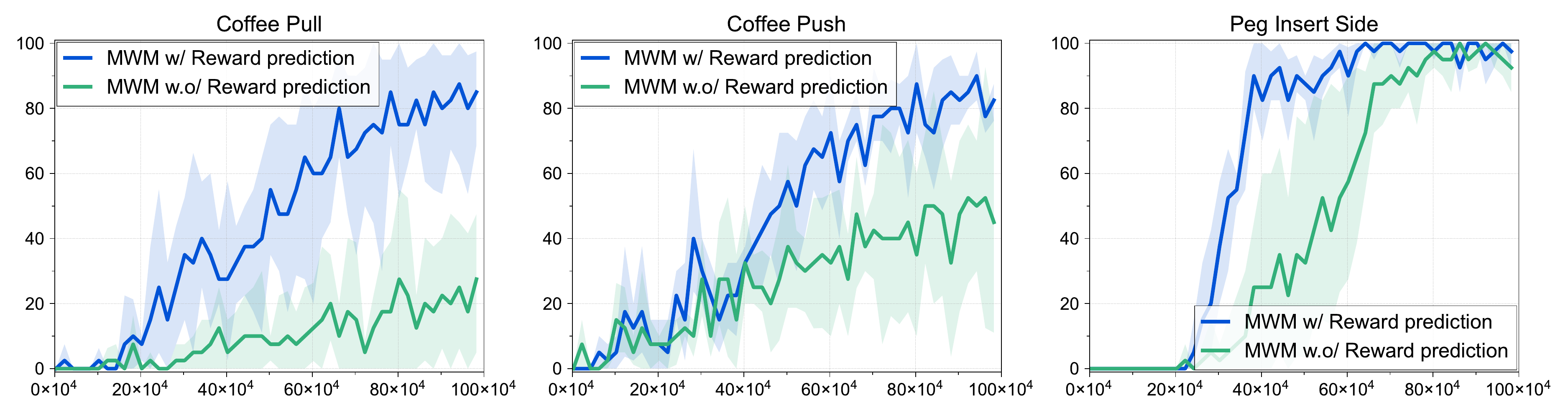}
    \includegraphics[width=0.96\textwidth]{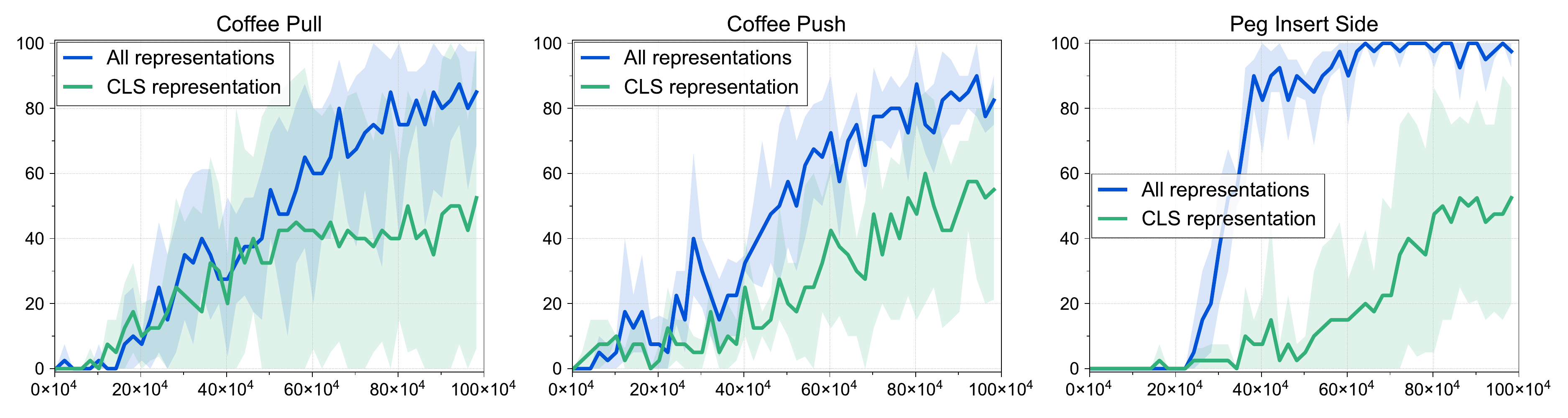}
    \includegraphics[width=0.96\textwidth]{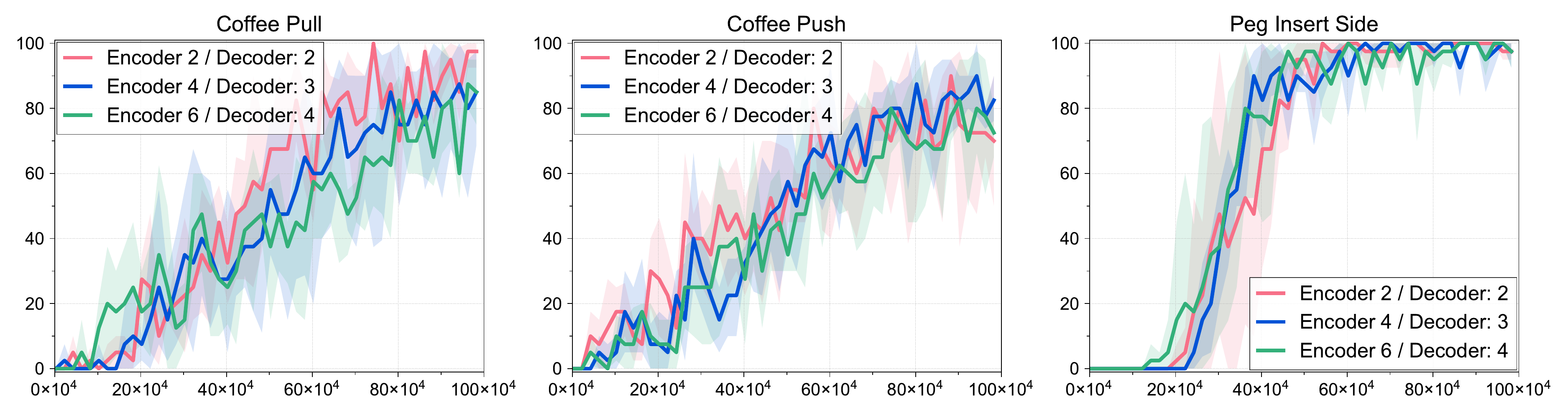}
    \includegraphics[width=0.96\textwidth]{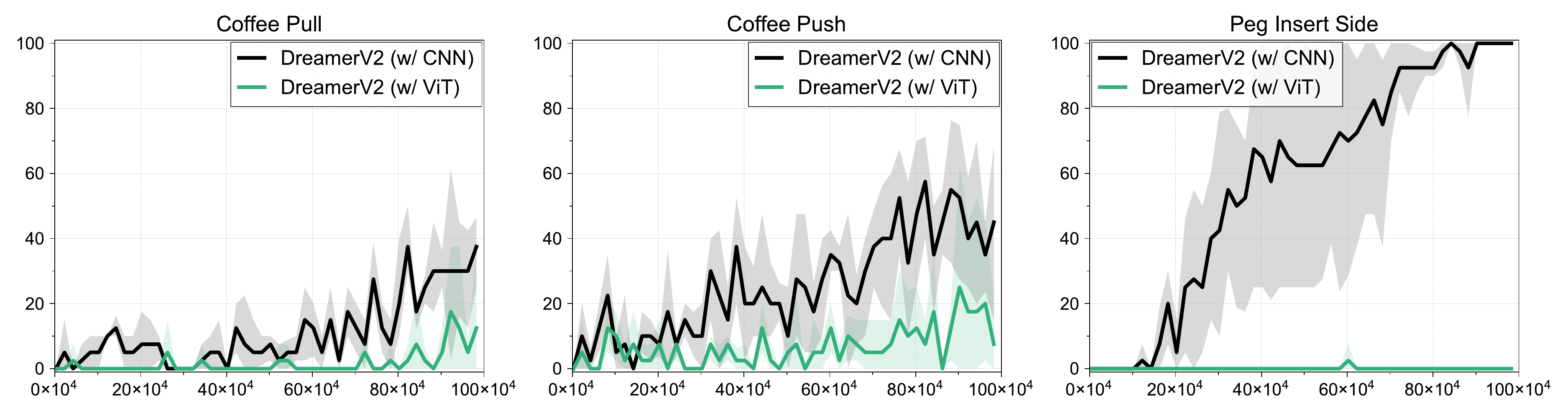}
    \caption{
    Learning curves on individual tasks used in ablation studies and analysis.
    The solid line and shaded regions represent the mean and bootstrap confidence interval across 4 runs.
    }
    \label{fig:metaworld_ablation_all}
\end{figure*}

We provide additional analysis in~\cref{fig:metaworld_additional_ablation} and learning curves on individual tasks in~\cref{fig:metaworld_ablation_all}.

\paragraph{Utilizing only CLS representation}
We ablate our design choice of utilizing all representations of $z_{t}^{c,0}$ for dynamics learning, instead of using only CLS representation as in MAE.
As shown in~\cref{fig:metaworld_ablation_class_token}, we find that utilizing all representations (\textit{i.e.,} CLS + Conv) outperforms the baseline (\textit{i.e.,} CLS), by encouraging the model to learn spatial information included in all representations.

\paragraph{Model size}
We report the performance of our method with varying number of layers for autoencoders in~\cref{fig:metaworld_ablation_model_size}.
We find that there are no significant differences between three models sizes we consider, which might be because Meta-world visual observations are not too complex. It would be interesting to investigate the effect of model sizes in more complex environments.

\paragraph{DreamerV2 with ViT}
In order to demonstrate that performance gain from our approach does not solely come from employing ViT instead of CNN,
we evaluate DreamerV2~w/~ViT that replaces CNN encoder and decoder with ViT encoder and decoder,  in~\cref{fig:metaworld_ablation_vit}. We find DreamerV2~w/~ViT exhibits severe instability during training, and sometimes becomes completely unable to solve the tasks.
We conjecture this might be because ViT suffers from unstable training without large data and regularization~\citep{dosovitskiy2020image}, which makes it difficult to learn world models end-to-end.

\clearpage

\clearpage

\section{Additional Meta-world Experiments with Longer Environment Steps}
\label{appendix:experiments_with_longer_steps}
\begin{figure*} [h] \centering
    \vspace{-0.2in}
    \subfigure[Hand Insert]
    {
    \includegraphics[width=0.3\textwidth]{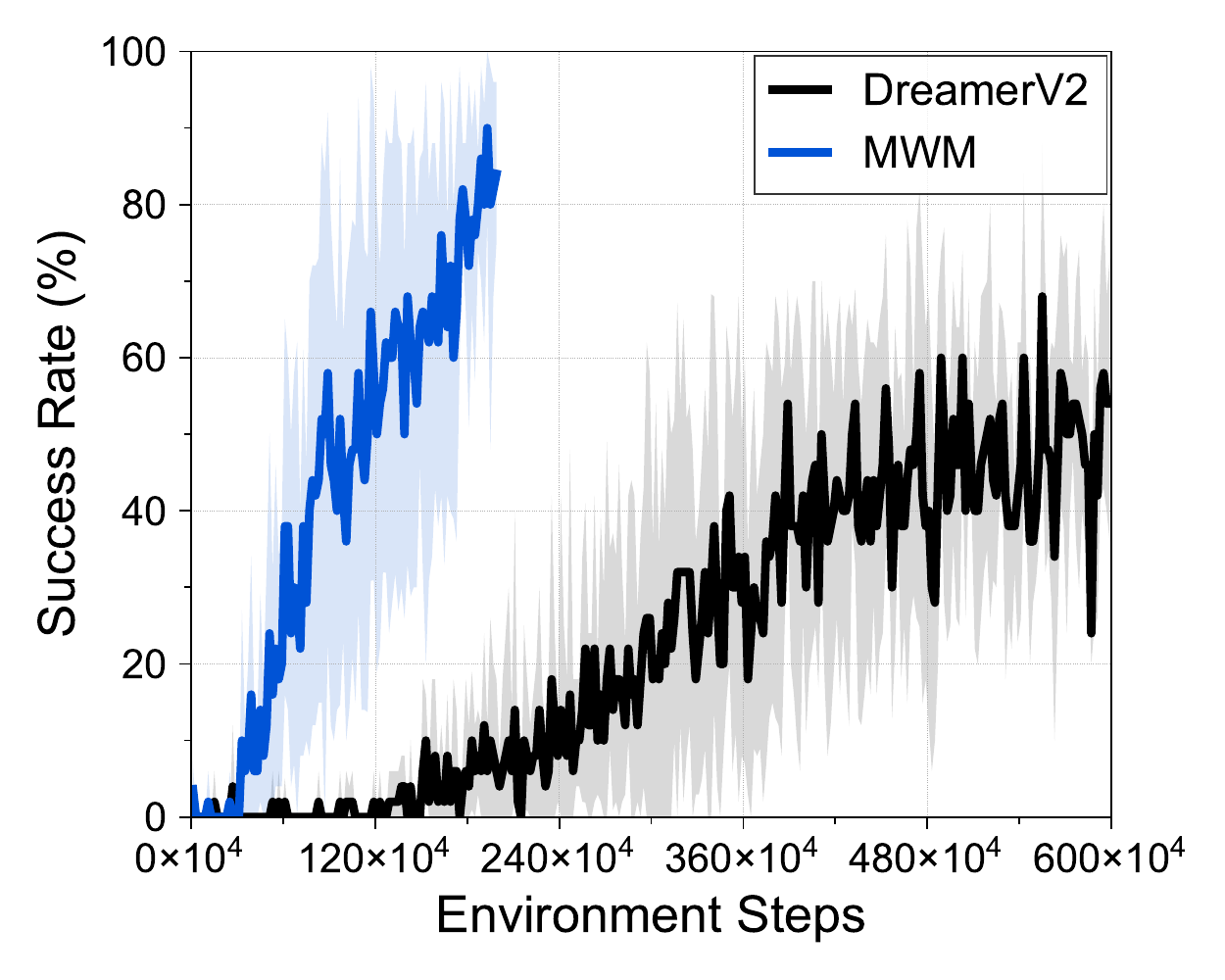}
    \label{fig:metaworld_long_hand_insert}}
    \subfigure[Shelf Place]
    {
    \includegraphics[width=0.3\textwidth]{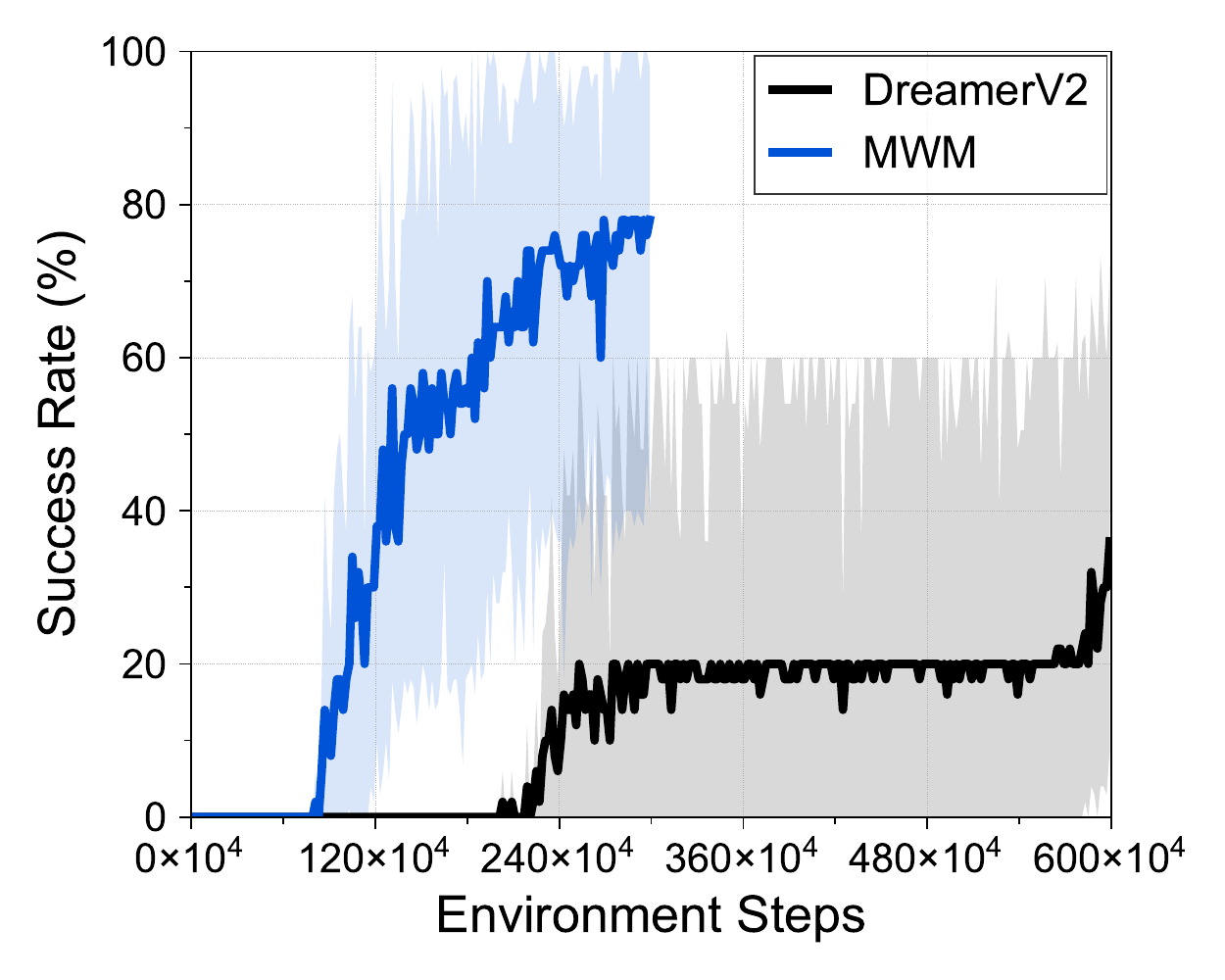}
    \label{fig:metaworld_long_shelf_place}}
    \vspace{-0.05in}
    \caption{
    Learning curves on (a) Hand Insert and (b) Shelf Place tasks from Meta-world as measured on the success rate. The solid line and shaded regions represent the mean and bootstrap confidence intervals, respectively, across five runs.
    }
    \label{fig:metaworld_long}
    \vspace{-0.1in}
\end{figure*}

We provide additional experiments that investigate the performance of DreamerV2 and MWM with more samples in~\cref{fig:metaworld_long}. Specifically, we report the performance of DreamerV2 over 6M environment steps, and find that DreamerV2 cannot outperform MWM even with much larger number of samples. This shows our approach allows for solving the tasks in a sample-efficient way with better asymptotic performance.

\section{Comparison with Additional Baselines}
\label{appendix:experiments_with_vae}
\begin{figure*} [h] \centering
    \vspace{-0.1in}
    \subfigure[Reach]
    {
    \includegraphics[width=0.315\textwidth]{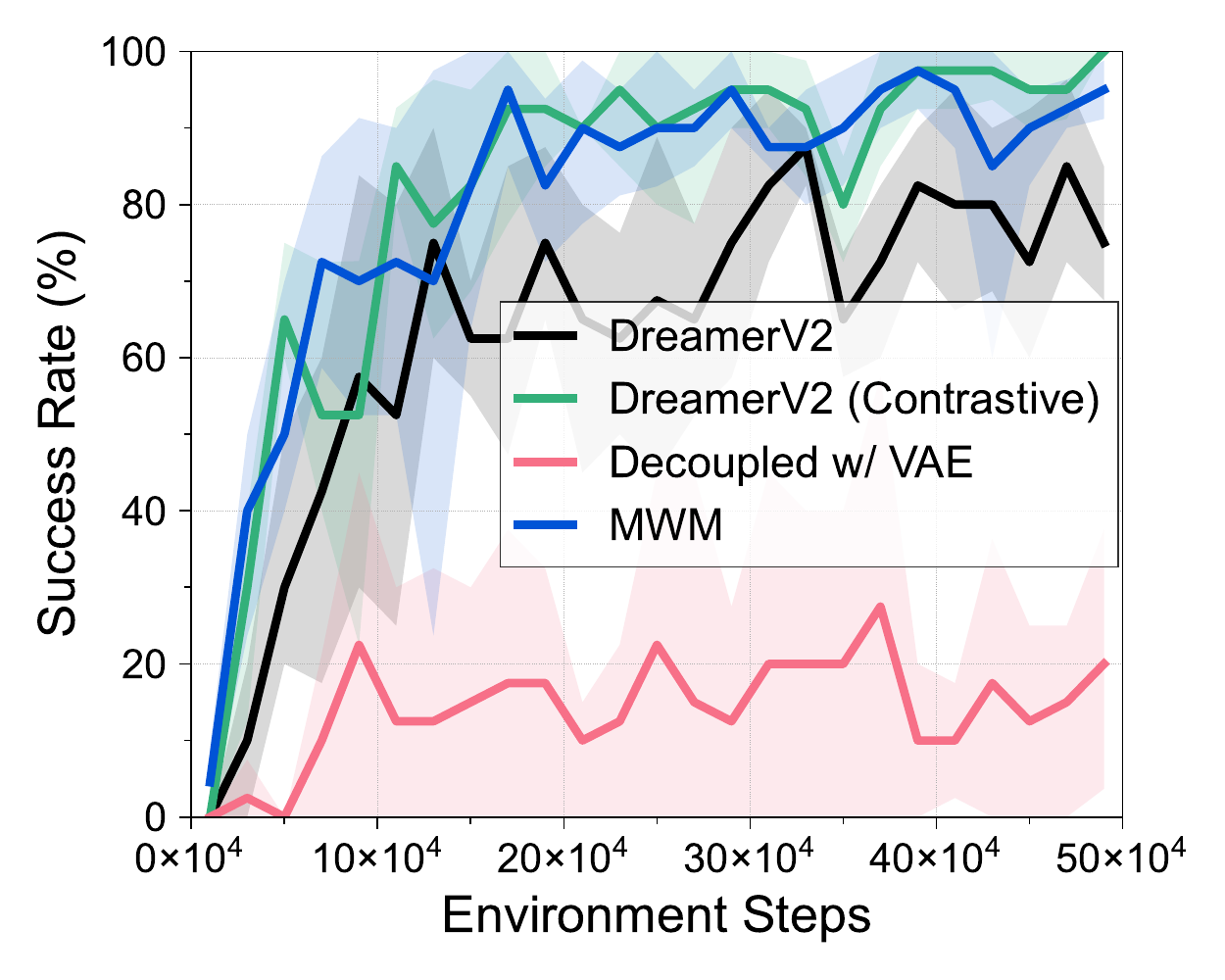}
    \label{fig:metaworld_reach_vae}}
    \subfigure[Push]
    {
    \includegraphics[width=0.315\textwidth]{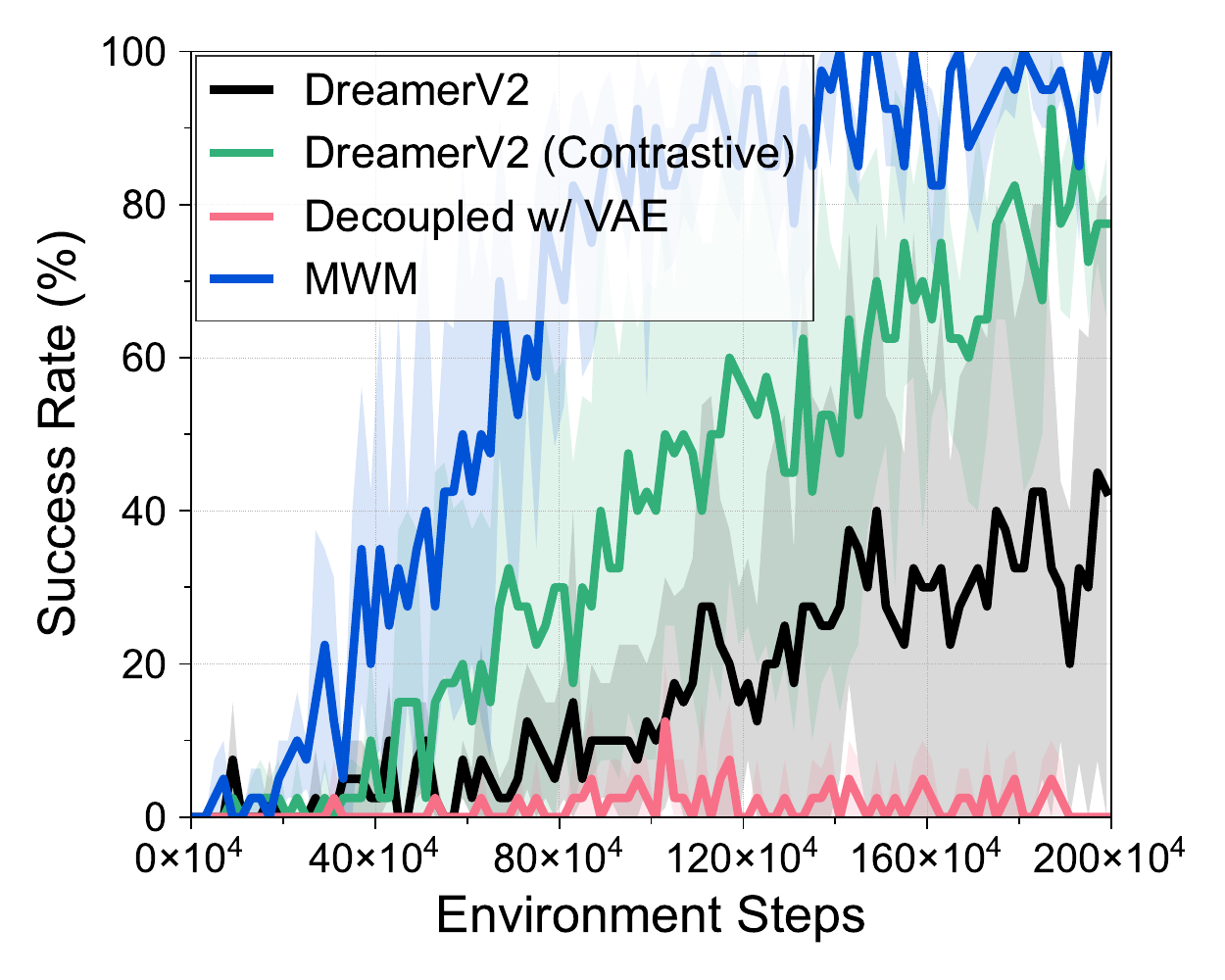}
    \label{fig:metaworld_push_vae}}
    \subfigure[Pick Place]
    {
    \includegraphics[width=0.315\textwidth]{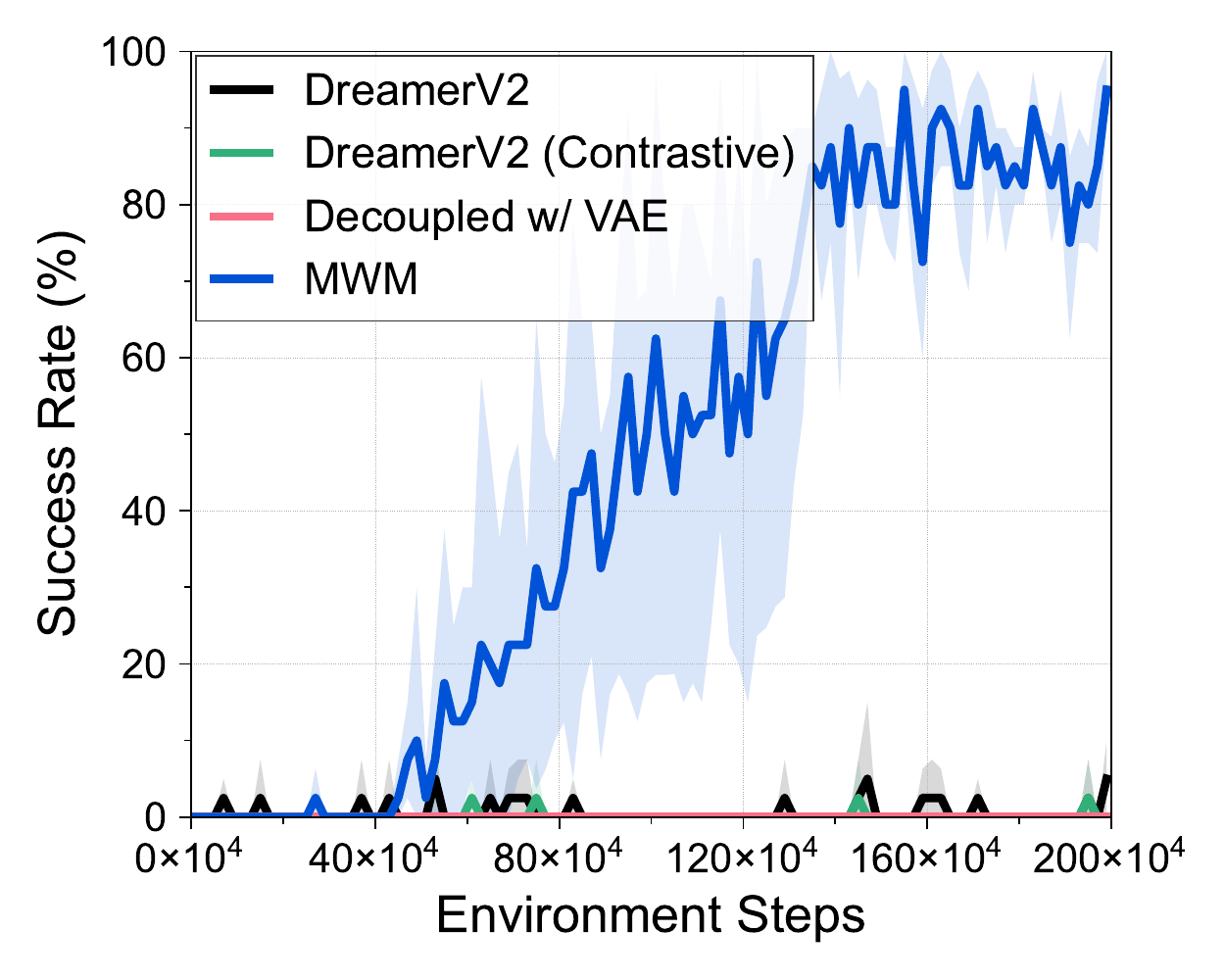}
    \label{fig:metaworld_pick_place_vae}}
    \vspace{-0.05in}
    \caption{
    Learning curves on (a) Reach, (b) Push, and (c) Pick Place tasks from Meta-world as measured on the success rate. The solid line and shaded regions represent the mean and bootstrap confidence intervals, respectively, across four runs.
    }
    \label{fig:metaworld_vae}
\end{figure*}

In this section, we provide additional experiments that compare MWM with additional baselines. Specifically, we compare MWM against (i) a baseline that decouples visual representation and dynamics learning by learning the model on top of frozen VAE~\citep{kingma2013auto} representations and (ii) a baseline that learns representations via contrastive learning~\citep{hafner2019learning} instead of reconstruction in~\cref{fig:metaworld_vae}.

\paragraph{Comparison with VAE} We find that the decoupled baseline based on VAE fails to solve most of the tasks, while MWM solves the target tasks. This shows that our representation learning method is crucial for performance improvement from the decoupling approach.
It would be an interesting investigation to consider additional baselines that extracts keypoints and learns a dynamics model on top of them~\citep{minderer2019unsupervised,manuelli2020keypoints,lambeta2020digit,das2021model}.

\paragraph{Comparison with contrastive baseline}
Moreover, we also observe that DreamerV2 with contrastive representation learning outperforms DreamerV2 with reconstruction, which shows that contrastive learning could be better in capturing fine-grained details from small objects. But importantly we find MWM still outperforms DreamerV2 (Contrastive) especially on Pick Place task, which demonstrates the effectiveness of our decoupled approach for solving challenging manipulation tasks.

\clearpage

\section{Generalizability of Representations Learned with Reward Prediction}
\label{appendix:generalizability}

As we mentioned in~\cref{sec:discussion}, representation learning only with task-irrelevant information would be an interesting direction to further improve the applicability of our approach to diverse setups.
Nevertheless, we remark that reward prediction on a specific task can also encourage visual representations to capture task-irrelevant information useful for solving various manipulation tasks.
For instance, rewards are often designed to contain information about robot arm movements (\textit{e.g.,} translations and rotations) and fine-grained details about objects, which can be shared across various tasks.

\begin{figure*} [h] \centering
    \vspace{-0.1in}
    \subfigure[Pick Place]
    {
    \includegraphics[width=0.315\textwidth]{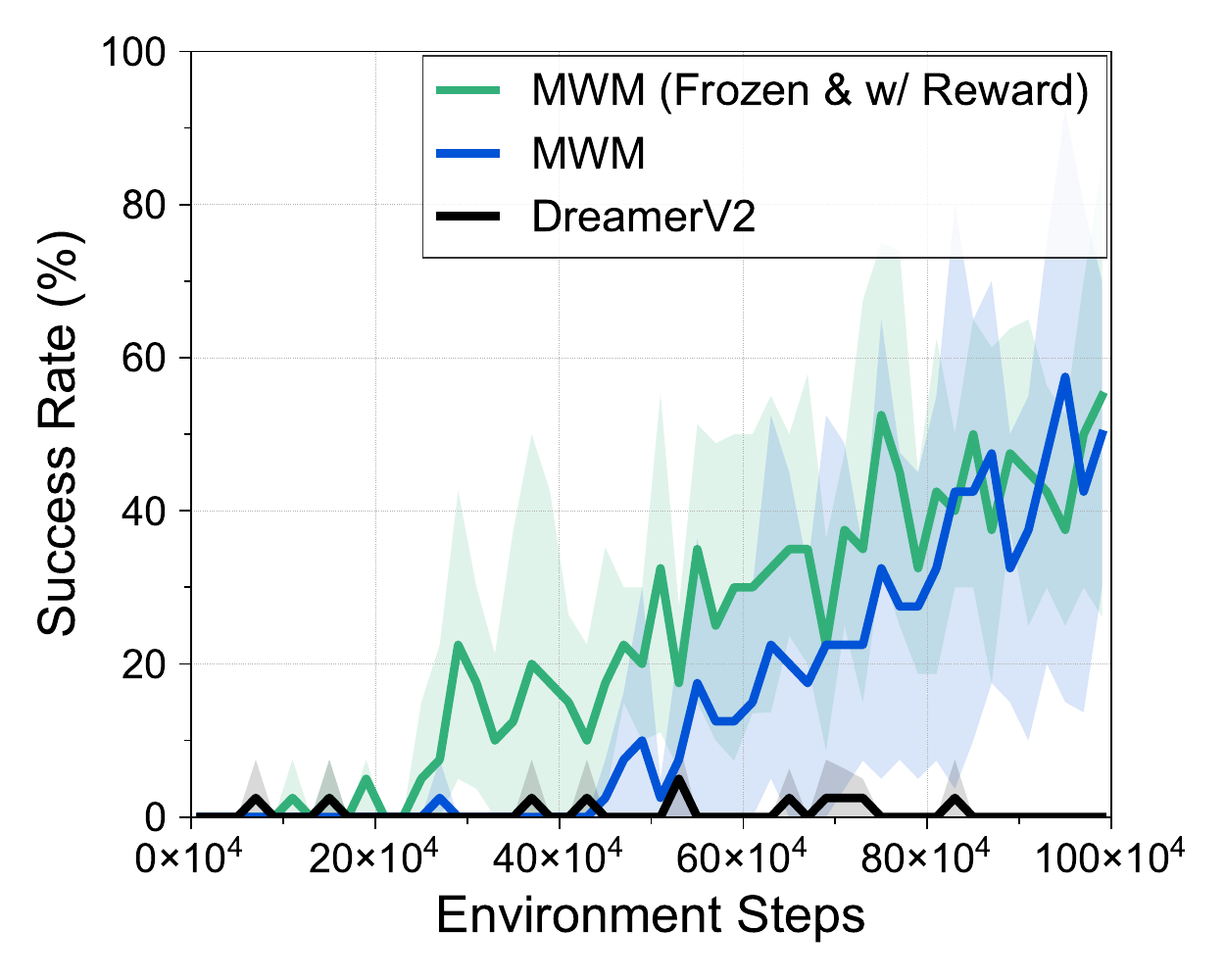}
    \label{fig:metaworld_pick_place_frozen}}
    \subfigure[Push Back]
    {
    \includegraphics[width=0.315\textwidth]{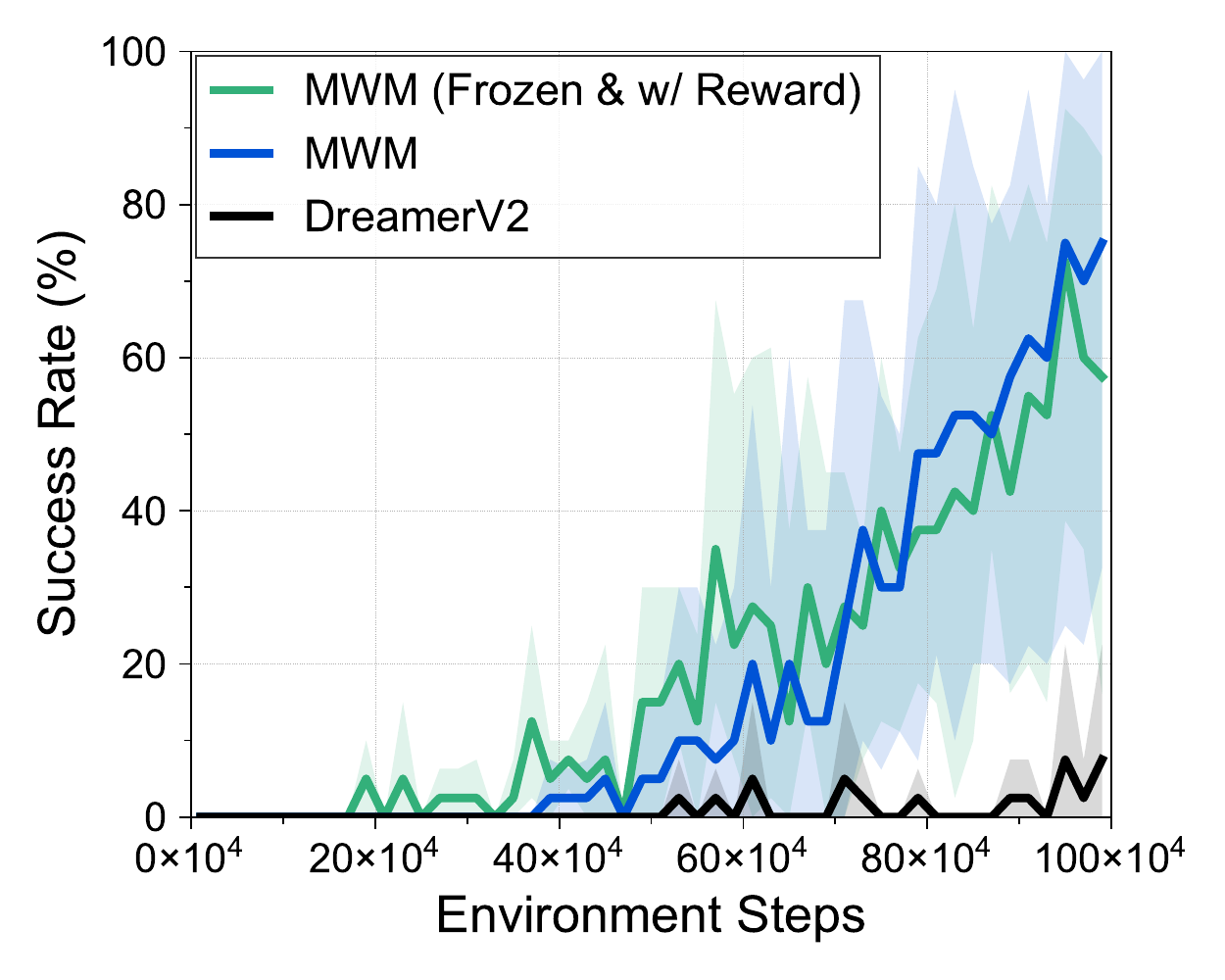}
    \label{fig:metaworld_push_back_frozen}}
    \subfigure[Drawer Open]
    {
    \includegraphics[width=0.315\textwidth]{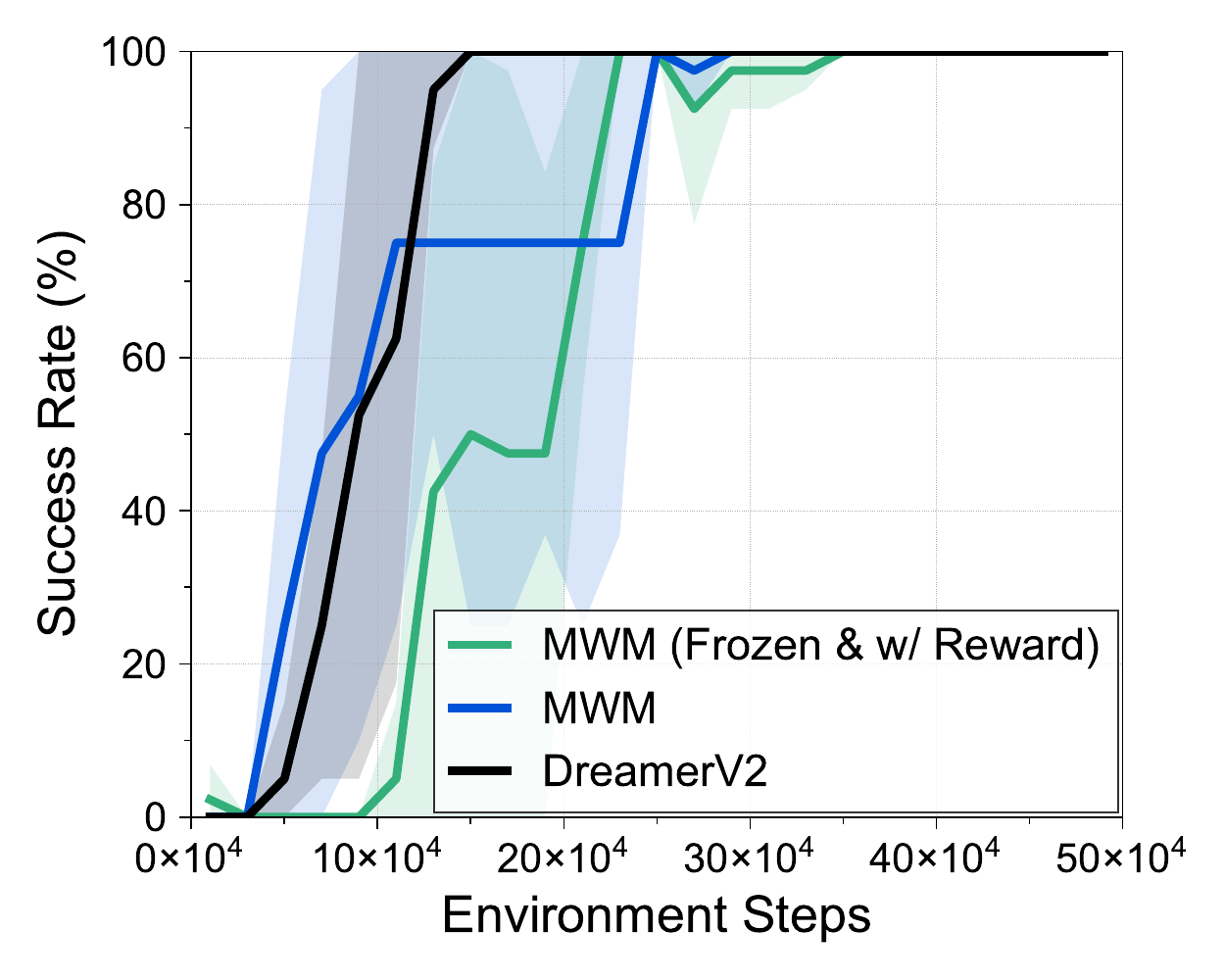}
    \label{fig:metaworld_drawer_open_frozen}}
    \vspace{-0.05in}
    \caption{
    Learning curves on (a) Pick Place, (b) Push Back, and (c) Drawer Open tasks from Meta-world as measured on the success rate. The solid line and shaded regions represent the mean and bootstrap confidence intervals, respectively, across four runs.
    }
    \label{fig:metaworld_frozen}
\end{figure*}

\paragraph{Transfer to unseen tasks} In order to empirically support this, we report the experimental results where we utilize frozen representations trained on Push task for solving manipulation tasks with a difference to push task: (i) Push Back that requires moving a block to a different position, (ii) Pick Place that requires picking up the block, and (iii) Drawer Open that contains unseen drawer object in the observation.
As shown in~\cref{fig:metaworld_frozen}, we observe that performance with frozen representations from unseen task can be similar to or better than the performance of MWM trained from scratch, which shows that representations learned with reward prediction can be versatile.

\paragraph{State regression analysis}

\begin{figure*} [h] \centering
    \vspace{-0.1in}
    \includegraphics[width=0.33\textwidth]{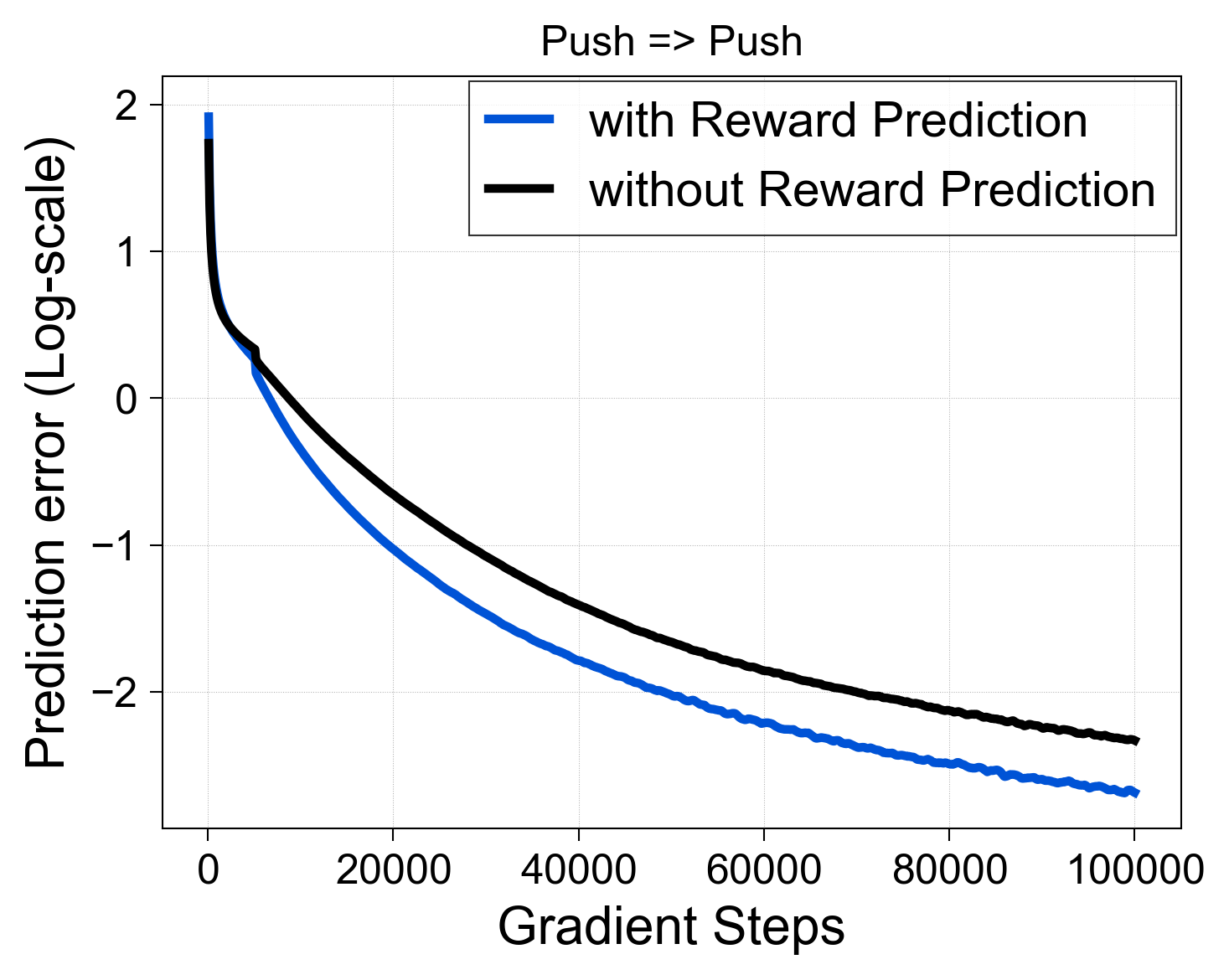}
    \label{fig:metaworld_state_regression_push}
    \includegraphics[width=0.33\textwidth]{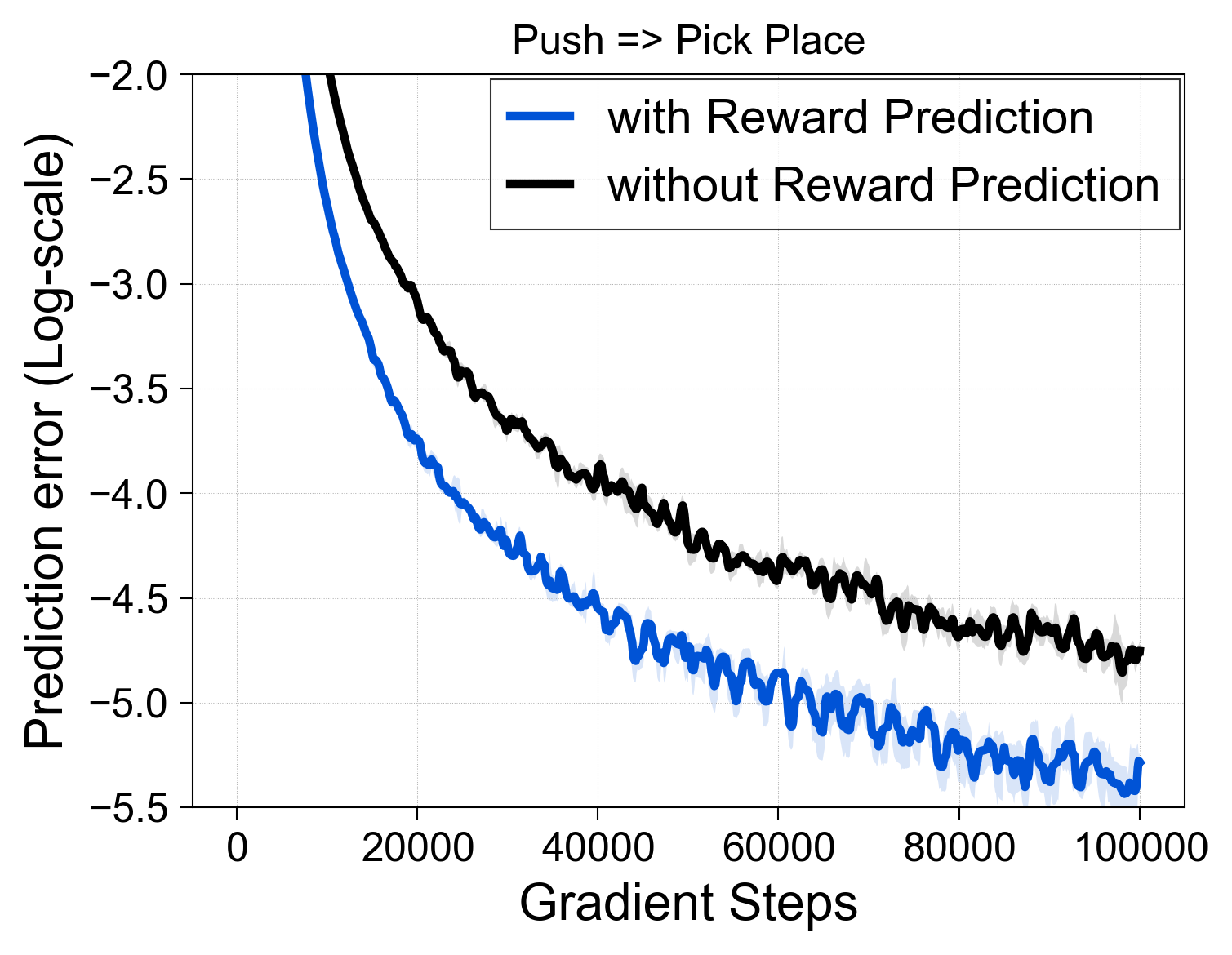}
    \label{fig:metaworld_state_regression_pick_place}
    \vspace{-0.05in}
    \caption{
    Learning curves of regression models that predicts proprioceptive states using the frozen representations pre-trained from Push task on (a) Push and (b) Pick Place tasks. The solid line and shaded regions represent the mean and bootstrap confidence intervals, respectively, across four runs.
    }
    \label{fig:metaworld_state_regression}
\end{figure*}

To further investigate how reward prediction affects the quality of learned representations, we provide additional experiments where we train a regression model to predict proprioceptive states available from the simulator.
Specifically, we first train MWM on Meta-world Push task with and without reward prediction, and train a regression model to predict the states in Push (seen) and Pick-place (unseen) tasks on top of frozen autoencoder representations from Push task. 
In~\cref{fig:metaworld_state_regression}, we observe that representations trained with reward prediction consistently achieve low prediction error. This supports that reward prediction can indeed provide useful information for robotic manipulation.

\end{document}